%% file: main.tex
\documentclass[journal]{IEEEtran}
\input{sources}
\begin{document}

\title{Federated Learning Across Decentralized and Unshared Archives for Remote Sensing\\Image Classification}

\author{Barı\c{s} B\"{u}y\"{u}kta\c{s},
        Gencer Sumbul,~\IEEEmembership{Member,~IEEE},
        Beg\"{u}m Demir,~\IEEEmembership{Senior Member,~IEEE}%
        \thanks{Barı\c{s} B\"{u}y\"{u}kta\c{s} and Beg{\"u}m Demir are with the Faculty of Electrical Engineering and Computer Science, Technische Universit\"at Berlin, 10623 Berlin, Germany, also with the BIFOLD - Berlin Institute for the Foundations of Learning and Data, 10623 Berlin, Germany.
        Email: \mbox{baris.bueyuektas@tu-berlin.de}, \mbox{demir@tu-berlin.de}.
        
        Gencer Sumbul is with the Environmental Computational Science and Earth Observation Laboratory (ECEO), École Polytechnique Fédérale de Lausanne (EPFL), 1950 Sion, Switzerland (e-mail: \mbox{gencer.sumbul@epfl.ch}).}%
}

\maketitle

\begin{abstract}

Federated learning (FL) enables the collaboration of multiple deep learning models to learn from decentralized data archives (i.e., clients) without accessing data on clients. Although FL offers ample opportunities in knowledge discovery from distributed image archives, it is seldom considered in remote sensing (RS). In this paper, as a first time in RS, we present a comparative study of state-of-the-art FL algorithms for RS image classification problems. To this end, we initially provide a systematic review of the FL algorithms presented in the computer vision and machine learning communities. Then, we select several state-of-the-art FL algorithms based on their effectiveness with respect to training data heterogeneity across clients (known as non-IID data). After presenting an extensive overview of the selected algorithms, a theoretical comparison of the algorithms is conducted based on their: 1) local training complexity; 2) aggregation complexity; 3) learning efficiency; 4) communication cost; and 5) scalability in terms of number of clients. After the theoretical comparison, experimental analyses are presented to compare them under different decentralization scenarios. For the experimental analyses, we focus our attention on multi-label image classification problems in RS. Based on our comprehensive analyses, we finally derive a guideline for selecting suitable FL algorithms in RS. The code of this work is publicly available at \url{https://git.tu-berlin.de/rsim/FL-RS}.
\end{abstract}


\IEEEpeerreviewmaketitle

\section{Introduction}

As remote sensing (RS) technology continues to advance, RS data has become increasingly vast in size and distributed across decentralized image archives (i.e., clients). For knowledge discovery from RS image archives, deep learning (DL) has been proven its success among all the data-driven approaches. The use of DL generally requires complete access to training data while learning the DL model parameters~\cite{Lyu:2022}. However, training data in some clients may not be accessible to the public (i.e., unshared) due to commercial interests, privacy concerns or legal regulations. The commercial RS data providers may consider training data as a valuable asset, and thus often restrict public access to protect their proprietary information and business interests. In RS, privacy concerns can also be crucial across various use cases (e.g., crop monitoring, damage assessment, wildlife tracking etc.) that may limit public access to training data on clients. As an example, in the context of crop monitoring through RS images, farmers might be hesitant to share detailed information about their crops \cite{tuia2023artificial}, and thus may not allow sharing crop labels. For wildlife tracking with RS images \cite{cerrejon2021no}, training data can be unshared with the public to avoid disclosing the precise locations of endangered species or sensitive ecosystems. There may also be legal restrictions (e.g., privacy laws, national security concerns etc.) in place that prevents public access to sensitive information contained within the training data on clients \cite{zhang2022progress}. 

When there is no access to training data on clients, federated learning (FL) can be utilized for training DL models on clients (i.e., local models) and finding the optimal model parameters on a central server (i.e., global model). As one of the first attempts to achieve FL, the federated averaging (FedAvg) algorithm is introduced in~\cite{mcmahan2017communication} to learn a global model by iterative model averaging in four consecutive steps, which are explained in the following. First, the central server sends the initial global model to all clients. Second, in each client, this model is trained on the corresponding local data of the client, leading to one local model. Third, each client sends the parameters of its local model back to the central server. Fourth, the central server updates the global model by averaging the parameters of all local models (i.e., parameter aggregation). These steps are repeated several times (i.e., rounds) until the convergence of the global model. This strategy is illustrated in Fig. \ref{figure}.

\input{FL_figure}
When training data is distributed across clients, training data in different clients might be not independent and identically distributed (non-IID). This has been found to be one of the main factors, affecting the convergence of the considered global model \cite{xie2023fedkl}. In RS, there are three main reasons for such training data heterogeneity across clients. First, the distribution of training image labels may vary across clients~\cite{kairouz2021advances}. This is more evident if multiple labels are assigned to each training image. Second, different clients can hold significantly different amounts of training data. Third, training images associated with a same class may have different data distributions across clients that is known as concept shift \cite{li2021fedbn}. This is due to the fact that the RS images belonging to the same class are highly affected because of physical factors (e.g., soil moisture, vegetation etc.) and atmospheric conditions in RS. 

When the iterative model averaging strategy is applied through the FedAvg algorithm, each local model is updated towards its local optima, which can be different among clients due to non-IID training data. This may lead to the deviation of the global model from its global optima after parameter averaging, and thus impeding the convergence of the global model and reducing its generalization capability~\cite{Li:2023}. To address this problem, in recent years advanced FL algorithms have been introduced mainly in the computer vision (CV) and machine learning (ML) communities for image classification problems (see Section \ref{sec:related_work}). We would like to note that although FL offers ample opportunities for classification of RS images across distributed RS image archives, it is seldom considered in RS (see Section \ref{sec:related_work} for a detailed review). As a first time in RS, in this paper, we present a comparative study of state-of-the-art FL algorithms in the context of RS image classification. The main contributions of this paper are summarized as follows:
\begin{itemize}
    \item We present a systematic review of the state-of-the-art FL algorithms in the CV and ML communities, and select the most effective algorithms based on their effectiveness with respect to training data heterogeneity across clients. 
    \item Together with an extensive overview of the selected algorithms, we provide their theoretical comparison in terms of: 1) local training complexity; 2) aggregation complexity; 3) learning efficiency; 4) communication cost; and 5) scalability.
    \item We provide an extensive experimental comparison of the selected algorithms under different decentralization scenarios. As the classification task of the experimental analysis, we focus our attention on multi-label image classification problems in RS.
    \item We derive some guidelines for selecting suitable FL algorithms in RS based on the level of non-IID training data, aggregation complexity, and local training complexity.   
\end{itemize}
The remaining part of this paper is organized as follows. We provide a detailed review of existing FL algorithms in Section~\ref{sec:related_work}. Among the existing algorithms, we consider several FL algorithms to present their extensive overview in the context RS image classification in Section~\ref{sec:overview}. Then, a theoretical comparison of the selected algorithms is conducted in Section~\ref{sec:theoretic_comp}. After the theoretical comparison, experimental analysis of these algorithms are carried out in Section~\ref{sec:exp_comp}, while Section~\ref{sec:exp_setup} provides the design of experiments. Section~\ref{sec:conclusion} concludes our paper with some guidelines for selecting suitable FL algorithms in RS and future research directions. We would like to note that we make the code of our paper publicly available at \url{https://git.tu-berlin.de/rsim/FL-RS} that allows to utilize all the considered algorithms for RS images. 

\begin{table*}[t]
\renewcommand{\arraystretch}{1.2}
\setlength\tabcolsep{9pt}
\caption{A Categorization of the Advanced Federated Learning Algorithms Addressing Training Data Heterogeneity Across Clients.\\ERM Denotes Empirical Risk Minimization. MPA Denotes Model Parameter Averaging}
\centering
\begin{tabular}{@{}llcccccccc@{}}
\hline
\multicolumn{2}{@{}l}{\multirow{4}{*}{Algorithm}} & \multicolumn{3}{c}{Local Training Strategy} & & \multicolumn{4}{c}{Aggregation Strategy}\\ \cline{3-5}\cline{7-10}
&& \thead[c]{Standard\\ERM} & \thead[c]{ERM with\\Proximal Term} & \thead[c]{Gradient\\Adjustment} & & \thead[c]{Standard\\MPA} & \thead[c]{Weighted\\MPA} & \thead[c]{Personalized\\MPA} & \thead[c]{Knowledge\\Distillation}\\ \hline
\multicolumn{2}{@{}l}{FedAvg\cite{mcmahan2017communication}} & \cmark & \xmark & \xmark & & \cmark& \xmark& \xmark& \xmark \\ \hline
\multirow{7}{*}{\thead[l]{Local\\Training\\Focused}} & FedProx \cite{li2020federated}& \xmark & \cmark & \xmark & & \cmark& \xmark& \xmark& \xmark\\ 
& SCAFFOLD \cite{karimireddy2020scaffold}& \cmark& \xmark& \cmark & & \cmark& \xmark & \xmark& \xmark\\ 
& MOON \cite{li2021model}& \xmark & \cmark & \xmark & & \cmark& \xmark& \xmark& \xmark \\ 
 & FedDyn \cite{acar2021federated}& \xmark & \cmark & \xmark & & \cmark & \xmark& \xmark& \xmark\\ 
 & FedDC \cite{gao2022feddc}& \xmark& \cmark& \cmark & & \cmark& \xmark& \xmark& \xmark\\ 
 & FedAlign \cite{mendieta2022local}& \xmark& \cmark& \xmark & & \cmark & \xmark& \xmark& \xmark\\ \hline
\multirow{10}{*}{\makecell[l]{Model\\Aggregation\\Focused}} & FedNova \cite{wang2021novel}& \cmark & \xmark & \xmark & & \xmark& \cmark& \xmark& \xmark\\ 
& FedMA \cite{wang2020federated}& \cmark& \xmark & \xmark & & \xmark& \cmark& \xmark& \xmark\\ 
& FedBN \cite{li2021fedbn}& \cmark& \xmark& \xmark & & \xmark& \cmark& \xmark& \xmark\\ 
& Mlmg \cite{qin2021mlmg}& \cmark& \xmark & \xmark & & \xmark& \xmark& \cmark& \xmark\\ 
& pFedLA \cite{ma2022layer}& \cmark& \xmark& \xmark & & \xmark& \xmark& \cmark& \xmark\\ 
& FedDF \cite{lin2020ensemble}& \cmark& \xmark& \xmark & & \xmark& \xmark& \xmark& \cmark\\ 
& FedBE \cite{chen2020fedbe} & \cmark& \xmark& \xmark & & \xmark& \xmark& \xmark& \cmark\\ 
& FedAUX \cite{sattler2021fedaux}& \cmark& \xmark& \xmark & & \xmark& \xmark& \xmark& \cmark\\ 
& FedGen \cite{zhu2021data}& \cmark& \xmark& \xmark & & \xmark& \xmark& \xmark& \cmark\\ \hline
\end{tabular}
\label{algs}
\end{table*}
\section{Recent Advances in Federated Learning Addressing Training Data Heterogeneity\\ Across Clients}
\label{sec:related_work}
In this section, we initially survey the advanced FL algorithms in the CV and ML communities, and then present the recent advances in FL from decentralized RS archives. We focus our attention on the FL algorithms, aiming to alleviate the effects of training data heterogeneity across clients during FL training.

\subsection{Recent Advances in Federated Learning for Computer Vision and Machine Learning}
We categorize the existing FL algorithms recently proposed in the CV and ML communities into two main groups: 1) local training focused algorithms~\cite{li2020federated,karimireddy2020scaffold,li2021model,acar2021federated,gao2022feddc,mendieta2022local}; and 2) model aggregation focused algorithms~\cite{wang2021novel,wang2020federated,li2021fedbn,qin2021mlmg,ma2022layer,lin2020ensemble,chen2020fedbe,sattler2021fedaux,zhu2021data} based on their advanced local training and aggregation strategies during FL that is summarized in Table \ref{algs}. 

Under the first category, FL algorithms enhance standard empirical risk minimization of local client training by: i) adding auxiliary loss functions (i.e., proximal terms)~\cite{li2020federated,li2021model,acar2021federated,gao2022feddc,mendieta2022local}; and ii) utilizing gradient adjustment techniques~\cite{karimireddy2020scaffold,gao2022feddc}. In FL, adding proximal terms to the local training of clients is found to be successful to limit local models deviating from the global model. As an example, in~\cite{li2020federated} a federated optimization algorithm (denoted as FedProx) is introduced to include a proximal term in each local training based on the $l_2$ distance between the parameters of the global model and each local model. In \cite{li2021model}, a model-contrastive federated learning algorithm (denoted as MOON) is introduced to utilize NT-Xent loss function for reducing the distance of feature distributions among clients while increasing that of among different local rounds. In \cite{acar2021federated}, the federated dynamic regularizer (FedDyn) algorithm is proposed to include a regularization term in local client training for aligning the model parameters between the central server and each client. It is worth noting that proximal terms may impede the convergence of local models as they constrain local updates. As an alternative, gradient adjustment techniques are also employed in FL for limiting local models deviating from the global model. For example, in \cite{karimireddy2020scaffold}, the stochastic controlled averaging algorithm (denoted as SCAFFOLD) is introduced to adjust the local gradient updates in clients based on the differences between the update directions of the global model and each local model. In \cite{mendieta2022local}, a distillation-based regularization method (denoted as FedAlign) is proposed to enhance the generalization capability of local models rather than forcing them to be closer to the global model as in the above-mentioned algorithms. To this end, it regularizes the Liptschitz constants of the considered neural network blocks through a proximal term. In \cite{gao2022feddc}, a federated learning algorithm with local drift decoupling and correction (FedDC) is introduced to combine both strategies, adding proximal terms and utilizing gradient adjustment techniques. To this end, in local client training, it defines a proximal term based on an auxiliary local drift variable (which tracks the gap between the model parameters of the client and the global model) in addition to applying gradient correction similar to the SCAFFOLD algorithm. 

Under the second category of FL algorithms, existing studies mainly aim to improve the standard parameter aggregation of iterative model averaging through: i) weighted model averaging; ii) personalized model averaging; and iii) knowledge distillation. As an example of weighted model averaging, in \cite{wang2021novel}, the federated normalized averaging (FedNova) algorithm is proposed to reduce inconsistencies among local models due to training data heterogeneity across clients. To achieve this, it normalizes the parameters of local models before applying aggregation on the central server. In \cite{wang2020federated}, the federated matched averaging (FedMA) algorithm is proposed for layer-wise weighting the parameters of local models during aggregation, while parameter weights are found through a bipartite matching optimization problem. In \cite{hsu2019measuring}, the federated averaging with server momentum (denoted as FedAvgM) algorithm is proposed to mitigate the differences among local models by averaging local model parameters based on the momentum of their changes throughout FL rounds. A federated learning algorithm (denoted as FedBN) is introduced in \cite{li2021fedbn} to alleviate concept shift between clients by excluding batch normalization parameters from the aggregation stage of FL. It is noted that in FedNova, FedMA and FedBN algorithms, a single global model is exploited on the central server. However, obtaining a single global model may not be sufficient for accurate FL when data heterogeneity across clients is high. To address this, personalized model averaging is found to be successful for enhancing cooperation among comparable clients when training data is non-IID among clients. As an example for such strategy, clustering clients and then employing one global model for each cluster is proposed in the multi-local and multi-global model aggregation (Mlmg) algorithm \cite{qin2021mlmg}. In \cite{ma2022layer}, the layer-wised personalized federated learning algorithm (denoted as pFedLA) is introduced to characterize the importance of each layer from different clients, and thus to optimize the personalized model aggregation for clients with heterogeneous data by using hypernetworks. Most of the FL algorithms, which are based on both weighted and personalized model averaging, require clients and the central server to employ a same DL model. To overcome this limitation, knowledge distillation based aggregation strategies have been recently proposed in FL. These strategies aim to obtain the global model by utilizing aggregated knowledge from clients rather than directly aggregating their model parameters. It is achieved by training a global model (which generally has a less number of parameters than local models) on the outputs of local models that requires clients to send their outputs rather than parameters. Main advantage of such strategies is to reduce the communication cost of FL compared to the parameter averaging based aggregation strategies. They also indirectly enhance the effectiveness of the global model towards training data heterogeneity due to the elimination of direct parameter averaging of local models. As an example for such algorithms, in \cite{lin2020ensemble} the federated distillation fusion (FedDF) algorithm is proposed to learn the global model as a student model by distilling the ensemble of clients as teacher models based on the average of their model outputs. Differently than FedDF, the federated bayesian ensemble (FedBE) algorithm is introduced in \cite{chen2020fedbe} to employ different importances of clients for knowledge distilation through Bayesian model ensemble. Sattler et al. \cite{sattler2021fedaux} proposes a FL algorithm (denoted as FedAUX) to define a certainity score of each client based on the pre-training of local models for weighting ensemble predictions during knowledge distillation. FedDF, FedBE and FedAUX algorithms require the availability of an unlabeled auxiliary training set or synthetically generated data, which is accessible to all clients during training. This may not be always feasible. To eliminate this requirement for knowledge distillation based parameter aggregation, in \cite{zhu2021data} an algorithm for federated distillation via generative learning (FedGen) is proposed. In this algorithm, a lightweight generator is learned in the central server to aggregate information from clients without the need for using an auxiliary data. 

\subsection{Recent Advances in Federated Learning for Remote Sensing}
As it can be seen from our survey, the development of FL algorithms has been widely studied in the CV and ML communities. However, the development of FL algorithms in RS has been attracted attention only in recent years although it offers ample opportunities for many RS problems. Unlike ML and CV, in RS there are only few FL studies, addressing training data heterogeneity across clients \mbox{\cite{zhang2023federated,zhang2022prototype,Buyuktas:2023}}. In {\cite{zhang2023federated}}, a FL scheme with prototype matching (FedPM) is proposed for pixel-based RS image classification to reduce data distribution divergence across heterogeneous RS image data on clients. To this end, in FedPM prototypical representation of each class is first obtained on each client, and then aggregated on the central server. Such prototypical representations are used in a proximal term to regularize local training and to reduce feature distribution differences across clients. In {\cite{zhang2022prototype}}, a FL algorithm combined with prototype-based hierarchical clustering (Fed-PHC) is introduced for pixel-based RS image classification to alleviate the limitations of training data heterogeneity through personalized model averaging. To this end, in Fed-PHC clients are clustered based on the training data distribution similarities across clients. Then, one global model is employed for each cluster similar to the Mlmg algorithm {\cite{qin2021mlmg}}. In {\cite{Buyuktas:2023}}, a multi-modal FL framework is proposed to learn from decentralized multi-modal RS image archives in the context of RS image classification. To this end, this framework operates FL through iterative model averaging when clients are associated with different RS image data modalities by defining modality-specific backbones. For addressing training data heterogeneity across clients and modalities, this framework models mutual information between distinct modalities and exploits batch whitening in place of batch normalization. In {\cite{li2024towards}}, a multiparty personalized collaborative learning (MPCL) framework is introduced. The MPCL framework aims to employ customized global models specific to different RS data characteristics and tasks by decoupling personalized model optimization from global model learning. In {\cite{wang2024personalized}}, a personalized adaptive distillation (PAD) scheme is proposed for the model parameter aggregation stage of FL in the context of few-shot classification problems in RS. The PAD scheme aims to prevent local model overfitting during FL training by employing adaptive weight initialization during parameter aggregation. This is achieved by retaining the shallower layers of the local models, while personalizing the deeper layers with adaptive aggregation weights. This leads to an increase in the generalization capabilities of local models, while addressing the limitations of training data heterogeneity.

In addition to training data heterogeneity, communication cost and privacy has recently emerged as other research directions for FL in RS. As an example, in {\cite{chen2024free}}, a privacy-reserving RS target fine-grained classification framework based on FL (which is denoted as PRFL) is introduced. PRFL aims to reduce communication cost among clients and a central server during FL training in the context of RS image scene classification. This is achieved through dynamic parameter decomposition, which is a low-rank matrix decomposition strategy for compressing DL model parameters. In {\cite{zhang2023local}}, local differential privacy embedded FL algorithm (denoted as LDP-Fed) is proposed to protect data and DL model privacy from white-box membership inference attacks for scene classification problems in RS. To this end, in LDP-Fed local differential privacy perturbation is applied to DL model parameters for preventing to obtain original local models from clients. In {\cite{zhu2023privacy}}, blockchain-empowered privacy-preserving FL algorithm is proposed to protect the privacy of DL models from model poisoning attacks in the context of RS image scene classification. 

\section{An Overview of the Selected\\ Federated Learning Algorithms}
\label{sec:overview}
In this section, we focus our attention on FedProx \cite{li2020federated}, SCAFFOLD \cite{karimireddy2020scaffold}, MOON \cite{li2021model}, FedDC \cite{gao2022feddc}, FedNova \cite{wang2021novel}, pFedLA \cite{ma2022layer}, and FedBN \cite{li2021fedbn} algorithms due to their proven success in FL and present their extensive overview for RS image classification problems. In detail, based on the effectiveness of these algorithms towards training data heterogeneity across clients, we categorize them into two main groups: 1) the local training focused algorithms; and 2) the model aggregation focused algorithms.

Let $K$ be the total number of clients and $C_i$ denotes the $i$th client for $1\leq i\leq K$. Let $D_i = \{(\boldsymbol{x}_{i}^z, \boldsymbol{y}_{i}^z)\}_{z=1}^{M_i}$ be the corresponding local training set of $C_i$, where $M_i$ is the number of images in $D_i$, $\boldsymbol{x}_{i}^z$ is the $z$th RS image in $D_i$ and $\boldsymbol{y}_{i}^z$ is the corresponding image annotation for $P$ unique classes. In the context of RS image classification, training image annotations can be given at scene-level or pixel-level. When the annotations are given at scene-level, each training image is annotated by either multi-labels or a single label (which is associated to the most significant content of the image). Let $\phi_i$ be the local DL model of $C_i$ and $w_i$ be the set of its parameters. $\phi_i$ provides the class probabilities $\boldsymbol{r}_{i}^{z}$ associated with $\boldsymbol{x}_{i}^z$ (i.e., $\phi_i(\boldsymbol{x}_{i}^{z};w_i)=\boldsymbol{r}_{i}^{z}$) while $z\in\{1,\ldots,M_i\}$. $\phi_i$ is locally trained on $D_i$ for $E$ epochs by using the cross entropy (CE) loss function $\mathcal{L}_{\text{CE}}$, which can be written as either categorical CE (for pixel-based and scene-level single-label image classification) or binary CE (for scene-level multi-label image classification). Then, for $C_i$ finding the optimum local model parameters $w_i^*$ by minimizing the corresponding local objective $\mathcal{O}_i$ (i.e., empirical risk minimization) can be formulated as follows:
\begin{equation}
\begin{aligned}
\mathcal{O}_i (\mathcal{B};w_i) &=\!\!\!\!\!\!\! \sum_{(\boldsymbol{x}_i^z,\boldsymbol{y}_i^z)\in \mathcal{B}}\!\!\!\!\mathcal{L}_{\text{CE}}(\phi_i(\boldsymbol{x}_{i}^z;w_i), \boldsymbol{y}_{i}^{z}),\\ 
w_i^* &= \arg \min_{w_i} \mathcal{O}_i(D_i;w_i).
\end{aligned}
\end{equation}
On decentralized and unshared training sets of the all clients $D=\bigcup_ {i \in \{ 1, 2,..., K \} } D_i$, FL algorithms aim to obtain the optimum parameters $w^*$ of a global model that minimizes all the local objectives as follows: 
\begin{equation}
w^* = \arg \min_{w} \sum_{i=1}^{K} \frac{\lvert D_i \rvert}{\lvert D \rvert} \mathcal{O}_{i}(D_i;w).
\label{minimization}
\end{equation}
To this end, the parameters of the global model are updated through the aggregation of local model parameters at each FL round as follows:
\begin{equation}
w = \sum_{i=1}^{K} \alpha_i w_i,
\label{fedavg}
\end{equation}
where $\alpha_i$ is a hyperparameter that adjusts the importance of the parameters of the client $C_i$ for aggregation. After updating the global model parameters, the parameters of each local model are updated based on $w$. Local training and parameter aggregation are consecutively repeated several times until the convergence of the global model (which is generally defined by the number of communication rounds $R$). 

When the standard iterative model averaging strategy (e.g., FedAvg algorithm) is applied in FL, each local model is updated towards its local optima based on the standard empirical risk minimization. The local optima of each client can be different from each other when training data is heterogeneous across clients (i.e., non-IID training data). This may lead to the deviation of the global model from its global optima after parameter averaging, since equal importance is given to each client in the FedAvg algorithm ($\alpha_i$ = $\lvert D_i \rvert / \lvert D \rvert $). 

\begin{algorithm}[t]
\SetAlgoLined
\Input{$\bigcup_ {i \in \{ 1, 2,..., K \} } D_i$, $R$, $E$, $\eta$, $K$}
\DontPrintSemicolon
\SetKwFunction{FAgg}{ModelAggregation}
\SetKwProg{PAgg}{function}{:}{end}
\SetKwFunction{FLocal}{LocalTraining}
\SetKwProg{PLocal}{function}{:}{end}

\PLocal{\FLocal{$D_i$,$w$,\textcolor{SCAFFOLD_c}{$v$},\textcolor{SCAFFOLD_c}{$v_i$},\textcolor{FedDC_c}{$h_i$}}}{
$w_i \leftarrow w$\;
\For(\Comment*[f]{$E$ epochs}){$e\gets1$ \KwTo $E$}{
  \For{$\mathcal{B} \in D_i$}{
        $\mathcal{O}_i (\mathcal{B};w_i) =\sum_{ \mathcal{B}} \mathcal{L}_{\text{CE}}(\boldsymbol{r}_{i}^z, \boldsymbol{y}_{i}^{z})$\;
        \textcolor{MOON_c}{$\mathcal{L}_{\text{MC}}(\mathcal{B};w_i)\!=\!\!-\!\!\sum_{\mathcal{B}}\text{log}(\!\frac{e^ { S( \boldsymbol{f}_i^z, \boldsymbol{f}_{i,g}^z ) \!/\! \tau  } }  {  e^ { S( \boldsymbol{f}_i^z, \boldsymbol{f}_{i,g}^z ) \!/\! \tau  } \!\!+ e^ { S( \boldsymbol{f}_{i,p}^z, \boldsymbol{f}_{i,g}^z) \!/\! \tau  }}\!)$}\;
        \textcolor{FedDC_c}{$P_i(h_i, w_i, w) = {\lVert h_i + w_i - w \rVert}^2$}\;
        \textcolor{FedProx_c}{$R_i(w,w_i)=\frac{\gamma}{2} \lvert\lvert w-w_i \rvert\rvert^2$}\;
        $\nabla_{w_i} \gets \nabla_{w_i} (\mathcal{O}_{i} + \textcolor{FedDC_c}{P_i} +  \textcolor{MOON_c}{\mathcal{L}_{\text{MC}}} + \textcolor{FedProx_c} {R} )$\;
        $w_i\leftarrow  w_i - \eta\left(\nabla_{w_i}+ \textcolor{SCAFFOLD_c}  {v-v_i}\right)$\;
    }
    \textcolor{SCAFFOLD_c}{$v_i \leftarrow v_i-v + \frac{w-w_i}{U\eta}$}\;
}
\textcolor{FedDC_c}{$h_i \leftarrow h_i + \Delta w_i$}\;
\KwRet $w_i$, \textcolor{FedDC_c}{$h_i$}, \textcolor{SCAFFOLD_c}{$v_i$} \;
}

$w^* \leftarrow w_0$ \Comment*{$w$ initialization}
\textcolor{SCAFFOLD_c}{$v, v_1,\ldots,v_K \leftarrow 0$} \;
\textcolor{FedDC_c}{$h_1,\ldots,h_K \leftarrow 0$} \;
\For(\Comment*[f]{$R$ rounds}){$r\gets1$ \KwTo $R$}{
    \For(\Comment*[f]{$K$ clients}){$i\gets1$ \KwTo $K$}{
        $w_i$, \textcolor{FedDC_c}{$h_i$}, \textcolor{SCAFFOLD_c}{$v_i$} $\leftarrow \FLocal(D_i,w^*,\textcolor{SCAFFOLD_c}{v},\textcolor{SCAFFOLD_c}{v_i},\textcolor{FedDC_c}{h_i})$ \;
    }
    $w^* \leftarrow \sum_{i=1}^{K} \frac{\lvert D_i \rvert} {\lvert D \rvert} (w_i + \textcolor{FedDC_c}{h_i})$ \;
    \textcolor{SCAFFOLD_c}{$v \leftarrow v + \frac{1}{K} \sum_{i=1}^{K} \Delta v_i$}\;

}
\KwRet $w^*$\;
\caption{The local training focused FL algorithms. The parts marked in brown, purple, cyan and orange represent changes only to \textcolor{FedProx_c}{FedProx}, \textcolor{SCAFFOLD_c}{SCAFFOLD}, \textcolor{MOON_c}{MOON} and \textcolor{FedDC_c}{FedDC}, respectively.
}
\label{alg:fed}
\end{algorithm}

\subsection{The Local Training Focused Algorithms}
\label{ltb}

This subsection presents the FedProx, MOON, FedDC and SCAFFOLD algorithms that aim to address the challenges of training data heterogeneity across clients by enhancing the standard empirical risk minimization during local training of FL. To this end, proximal terms are added to local objectives in the FedProx, MOON and FedDC algorithms, while gradient adjustment techniques are utilized during the update of local model parameters in the SCAFFOLD and FedDC algorithms. These FL algorithms are summarized in Algorithm \ref{alg:fed}, while their details are given below.

\subsubsection{FedProx \cite{li2020federated}} This algorithm aims at limiting local models to excessively deviate from the global model that is one of the main effects of training data heterogeneity across clients during federated training of the global model. This is achieved by adding a $L_2$ regularization term $R_i(w,w_i)=\frac{\gamma}{2} \lvert\lvert w-w_i \rvert\rvert^2$ to each local objective. The weight of the $L_2$ regularization term is adjusted with the $\gamma$ parameter. A high value of $\gamma$ allows to restrict local parameter updates based on local objectives, leading the parameters of local models to be closer to the global model parameters. A small value of $\gamma$ allows the local model parameters to be updated more based on the local objectives. Since FedProx reduces the difference of local model parameters between clients by restricting each local model with respect to the global model, it increases the generalization capacity of the local models. Due to this, the convergence of the local models does not significantly deviate from each other depending on $\gamma$ even though the data distribution differences between the clients are large. 

\subsubsection{MOON \cite{li2021model}} This algorithm aims to overcome the effects of training data heterogeneity across clients by preventing the image feature spaces of local models to deviate from that of the global model. This is achieved by increasing the similarity of the image features obtained from the global model and the local model, while decreasing the similarity of the features obtained from the local models of the consecutive FL rounds. To this end, for a given mini-batch $\mathcal{B}$, it adds the model-contrastive loss function $\mathcal{L}_{\text{MC}}$ to each local objective that is defined as follows:
\begin{equation} 
\mathcal{L}_{\text{MC}}(\mathcal{B};w_i)\!=-\!\!\!\sum_{\boldsymbol{x}_i\in \mathcal{B}}\!\! \text{log}\left(\!\frac{e^ { S( \boldsymbol{f}_i^z, \boldsymbol{f}_{i,g}^z ) / \tau  } }  {  e^ { S( \boldsymbol{f}_i^z, \boldsymbol{f}_{i,g}^z ) / \tau  } \!\!+\! e^ { S( \boldsymbol{f}_{i,p}^z, \boldsymbol{f}_{i,g}^z) / \tau  }}\!\right),
\label{moon}
\end{equation}
where $S(.,.)$ measures the cosine similarity, $\tau$ is a temperature parameter. $\boldsymbol{f}_i^z$, $\boldsymbol{f}_{i,p}^z$ and $\boldsymbol{f}_{i,g}^z$ are the image features of $\boldsymbol{x}_i^z$ obtained by using: i) the parameters of the local model; ii) the local model parameters from the previous FL round; and iii) the global model parameters, respectively.

\subsubsection{SCAFFOLD \cite{karimireddy2020scaffold}} This algorithm aims to alleviate the deviation in the updates of each local model that results in slow and unstable convergence (i.e., client-drift) when training data is distributed across heterogeneous clients. To this end, this algorithm defines one variable for each local and global model to estimate the change in the parameter updates of these models. Then, the defined parameters are utilized for adjusting gradients during local training. In detail, SCAFFOLD estimates the update direction of each client with a set of local parameters $v_i$ (i.e., the local control variate of $C_i$) and that of the global model with a set of global parameters $v$ (i.e., global control variate) based on the gradients of model parameters. Then, for a given mini-batch $\mathcal{B}$ the difference between $v_i$ and $v$ is added to the update rule of $w_i$ to adjust the corresponding gradients $\nabla_{w_i}$ as follows:   
\begin{equation}
w_i\leftarrow  w_i - \eta\left(\nabla_{w_i}\mathcal{O}_{i}(\mathcal{B};w_i)+v-v_i\right),
\label{scaffold}
\end{equation}
where $\eta$ is the learning rate. $\{v_i\}_{i=1}^K$ and $v$ are initially set to zero and updated after each FL round. For SCAFFOLD, two different options are defined to update the local control variate of each client. In the first option, for the $i$th client $v_i$ is updated based on the gradients of the global model parameters on the corresponding local training data as follows:
\begin{equation}
v_i \leftarrow \nabla_{w}\mathcal{O}_{i}(D_i;w).
\label{updates1}
\end{equation}
In the second option, the gradients of the previous FL rounds are utilized as follows:
\begin{equation}
v_i \leftarrow v_i-v + \frac{w-w_i}{U\eta}, 
\label{updates2}
\end{equation}
where $U$ is the number of local update steps. Since there is an additional forward pass over the local data in the first option, it is computationally more demanding than the second option. Thus, the second option is considered for the rest of our paper. After each local control variate is updated, $\{v_i\}_{i=1}^K$ is sent to the central server. The global control variate is updated at the central server as follows:
\begin{equation}
v \leftarrow v + \frac{1}{K} \sum_{i=1}^{K} \Delta v_i,
\label{updates3}
\end{equation}
where $\Delta v_i$ is the change of $v_i$ between two consecutive FL rounds.

\subsubsection{FedDC \cite{gao2022feddc}} This algorithm aims at preventing the effects of client-drift on FL of a global model by: i) adjusting gradients during local training through control variables as in the SCAFFOLD algorithm; and also ii) including a proximal term to the local objective. To this end, this algorithm defines a local drift variable $h_i$ for each client to estimate the change of local model parameters. $h_i$ is initially set to zero and updated after each local training of $C_i$ as follows:  
\begin{equation}
h_i \leftarrow h_i + \Delta w_i,
\label{driftt}
\end{equation}
where $\Delta w_i$ is the change of $w_i$ during local training. Then, in FedDC $h_i$ is exploited in the corresponding proximal term $P_i$ to control the change in the parameter updates of the global and local models. For given parameters of local and global models and the local drift variable, $P_i$ is defined as follows:
\begin{equation}
P_i (h_i, w_i, w) = {\lVert h_i + w_i - w \rVert}^2.
\label{penalized11}
\end{equation}
In addition to adding $P_i$ to the corresponding local objective of $C_i$, the local control variate $v_i$ and the global control variate $v$ are utilized for this algorithm as in SCAFFOLD. Accordingly, the update rule of $w_i$ is written as follows:
\begin{equation}
w_i\gets  w_i - \eta\left(\nabla_{w_i} [ \mathcal{O}_{i}(\mathcal{B};w_i)\! + \! P_i (h_i, w_i, w) ]\!+\!v\!-\!v_i\right),
\label{feddc11}
\end{equation}
where $v_i$ and $v$ are updated by using (\ref{updates2}) and (\ref{updates3}), respectively. After local training of each client is completed, the global model parameters are updated in the FedDC algorithm as follows:
\begin{equation}
w = \sum_{i=1}^{K} \frac{\lvert D_i \rvert}{\lvert D \rvert} (w_i + h_i).
\label{update13}
\end{equation}
\begin{algorithm}[t]
\SetAlgoLined
\Input{$\bigcup_ {i \in \{ 1, 2,..., K \} } D_i$, $R$, $E$, $\eta$, $K$}
\DontPrintSemicolon
\SetKwFunction{FAgg}{ModelAggregation}
\SetKwProg{PAgg}{function}{:}{end}
\SetKwFunction{FLocal}{LocalTraining}
\SetKwProg{PLocal}{function}{:}{end}

\PLocal{\FLocal{$D_i$,$w$}}{
$w_i \leftarrow w$\;
\For(\Comment*[f]{$E$ epochs}){$e\gets1$ \KwTo $E$}{
    \For{$\mathcal{B} \in D_i$}{
        $\mathcal{O}_i (\mathcal{B};w_i) =\sum_{ \mathcal{B}} \mathcal{L}_{\text{CE}}(\boldsymbol{r}_{i}^z, \boldsymbol{y}_{i}^{z})$\;
        $w_i\leftarrow  w_i - \eta\left(\nabla_{w_i}\mathcal{O}_{i}\right)$\;
    }
}
\KwRet $w_i$  \;
}

$w^*\!,\textcolor{pFedLA_c}{w_1^*,\ldots,w_K^*,w_{\lambda_1},\ldots,w_{\lambda_K}} \!\!\leftarrow\! w_0$ \Comment*{$w$ initialization}
\textcolor{pFedLA_c}{$\boldsymbol{t}_1,\ldots, \boldsymbol{t}_K \leftarrow \vec{\boldsymbol{0}}$}\; 
\For(\Comment*[f]{$R$ rounds}){$r\gets1$ \KwTo $R$}{
    \For(\Comment*[f]{$K$ clients}){$i\gets1$ \KwTo $K$}{
        $\textcolor{pFedLA_c}{\boldsymbol{\alpha}_i = \lambda_i(\boldsymbol{t}_i;w_{\lambda_i})}$ \;
        \textcolor{pFedLA_c}{$w_i^* = \bigcup_{l} \sum_{j=1}^K w_j^l \boldsymbol{\alpha}_{i,j}^l$} \;
        \textcolor{pFedLA_c}{$w^* \leftarrow w_i^*$}\;
        $w_i \leftarrow \FLocal(D_i,w^*)$ \;
        \textcolor{pFedLA_c}{$\boldsymbol{t}_i\leftarrow  \boldsymbol{t}_i - \eta(\nabla_{\boldsymbol{t}_i}w_i^*)^T\Delta w_i$}\;
        \textcolor{pFedLA_c}{$w_{\lambda_i}\leftarrow  w_{\lambda_i} - \eta(\nabla_{w_{\lambda_i}}w_i^*)^T\Delta w_i$}\;
    }
    $w^* \leftarrow \sum_{i=1}^{K} \frac{\lvert D_i \rvert} {\lvert D \rvert}(w_i \textcolor{FedBN_c}{\setminus \varphi_i})\textcolor{FedNova_c}{\frac{1}{E|D_i|/|\mathcal{B}|}}$\;
}
\textcolor{pFedLA_c}{$w^* \leftarrow w_1^*,\ldots,w_K^*$}\;
\KwRet $w^*$\;
\caption{The model aggregation focused FL algorithms. The parts marked in pink, magenta and blue represent changes only to \textcolor{FedNova_c}{FedNova}, \textcolor{FedBN_c}{FedBN} and \textcolor{pFedLA_c}{pFedLA}, respectively.}
\label{alg:fed_agg}
\end{algorithm}
\subsection{The Model Aggregation Focused Algorithms}

This subsection presents the FedNova, FedBN, and pFedLA algorithms that aim at alleviating the effects of training data
heterogeneity across clients by improving the standard parameter aggregation of iterative model averaging. To achieve this, the FedNova and FedBN algorithms utilize weighted parameter model averaging strategies, while the personalized global model aggregation of each client is exploited with hypernetworks in the pFedLA algorithm. These FL algorithms are summarized in Algorithm \ref{alg:fed_agg}, while their details are given below.

\subsubsection{FedNova \cite{wang2021novel}} During local training of FL, the considered clients may include different numbers of training samples (i.e., quantity skew) that is considered as one of the main reasons of training data heterogeneity across clients in FL. Quantity skew may lead the clients with higher numbers of training samples to have more impact on the global model update since the local update steps of these clients are higher than those of other clients. To eliminate this, the FedNova algorithm normalizes the parameters of local models according to the numbers of local updates before applying aggregation. Accordingly, at the end of each FL round it first weights the local model parameters of $C_i$ with the multiplicative inverse of the number of local updates (i.e., $ \frac{w_i}{E|D_i|/|\mathcal{B}|}$), and then sends the weighted local model parameters to the central server for parameter aggregation. 
\begin{table*}[t]
\renewcommand{\arraystretch}{1.1}
\setlength\tabcolsep{9pt}
\caption{The Theoretical Comparison of the Selected FL Algorithms. Three Marks "H" (High), "M" (Medium), or “L” (Low) Are Given for the Considered Criteria.}
\centering
\begin{tabular}{@{}llcccccc@{}}
\hline
Algorithm & & \thead[c]{Local Training\\Complexity} & \thead[c]{Aggregation\\Complexity} & \thead[c]{Learning\\Efficiency} & \thead[c]{Communication\\Cost} & Scalability \\ \hline
FedAvg \cite{mcmahan2017communication}  &  &  {L}  &  {L}   &  {M}    &   {L}      &     {L}        \\ \hline
\multirow{4}{*}{\thead[l]{Local\\Training\\Focused}} & FedProx \cite{li2020federated}         & {M}  & {L}  &  {M}  &  {L}   &     {H}          \\ 
& SCAFFOLD \cite{karimireddy2020scaffold}          &  {M}        &  {L}      &    {H}      &  {H}    &     {M}      \\ 
& MOON \cite{li2021model}          &   {H}   &   {L}  &  {H} &  {L}    &       {H}        \\ 
& FedDC \cite{gao2022feddc}           &    {M}   &  {L}   &   {M}   &  {H}       &      {H}       \\ \hline
\multirow{3}{*}{\thead[l]{Model\\Aggregation\\Focused}} & 
FedNova \cite{wang2021novel}           &   {L}   &  {M}    &  {L}   &  {L}    &     {L}         \\ 
& pFedLA \cite{ma2022layer}           & {L}  &  {H}    &   {L}   &  {L}          &      {L}       \\ 
& FedBN \cite{li2021fedbn}           &   {L}   &    {M}  &  {L}   &   {L}   &         {L}     \\ \hline
\end{tabular}
\label{theoretical_comp}
\end{table*}
\subsubsection{FedBN \cite{li2021fedbn}} This algorithm aims at preventing the effects of concept shift between clients by excluding batch normalization (BN) parameters from the aggregation stage of FL. To this end, BN parameters of each local model are not shared with the central server, and thus not included in parameter averaging of local models in the central server. Let $\varphi_i \in w_i$ be the set of BN parameters of the local model in $C_i$. The aggregation of local model parameters is achieved in the FedBN algorithm as follows:
\begin{equation}
w = \sum_{i=1}^{K} \frac{\lvert D_i \rvert}{\lvert D \rvert} (w_i \setminus \varphi_i). 
\label{fedbn}
\end{equation}

\subsubsection{pFedLA \cite{ma2022layer}} This algorithm aims at mitigating the effects of non-IID training data during FL by employing personalized global model aggregation for each client. This leads to learning one global model for each client (i.e., the set $\{w_1^*,\ldots,w_K^*\}$ of global models) instead of exploiting a single global model as in the other FL algorithms. Accordingly, the objective function of FL can be rewritten as follows:
\begin{equation}
w_1^*,w_2^*,...,w_K^* = \underset{\bigcup_{i}w_i}{\mathrm{argmin}} \sum_{i=1}^{K} \frac{\lvert D_i \rvert}{\lvert D \rvert} \mathcal{O}_{i}(D_i;w_i).
\label{minimization_p}
\end{equation}
To learn one global model for each client, the pFedLA algorithm aggregates the parameters of local models by considering the importance of each parameter differently for each client. To this end, it defines the aggregation weights $\boldsymbol{\alpha}_i =[\boldsymbol{\alpha}_i^{1},\boldsymbol{\alpha}_i^{2},...,\boldsymbol{\alpha}_i^{l}]$ of $C_i$, where $\boldsymbol{\alpha}_i^l$ is the aggregation weight vector of the $l$th parameter and $\boldsymbol{\alpha}_{i,j}^l$ is the corresponding aggregation weight associated with $C_j$ (i.e., $w_i^l$). To learn the aggregation weights of all the clients, in this algorithm hypernetworks~\cite{ha2016hypernetworks} are utilized at the central server. Let $\lambda_i$ be a hypernetwork of $C_i$ that takes the learnable embedding vector $\boldsymbol{t}_i$ and estimates $\boldsymbol{\alpha}_i$ as follows:
\begin{equation}
\label{hypernetwork1}
\boldsymbol{\alpha}_i = \lambda_i(\boldsymbol{t}_i;w_{\lambda_i}),
\end{equation}
where $w_{\lambda_i}$ is the parameters of the hypernetwork associated with $C_i$. Once the hypernetworks are included in (\ref{minimization_p}), the overall objective function of the pFedLA algorithm is written as follows:
\begin{equation}
\label{pfedla1}
\underset{\bigcup\limits_{i}\{\boldsymbol{t}_i, w_{\lambda_i}\!\}}{\mathrm{argmin}} \sum_{i=1}^{K} \frac{\lvert D_i \rvert}{\lvert D \rvert} \mathcal{O}_i\left(D_i; \bigcup_{l} \sum_{j=1}^K w_j^l \boldsymbol{\alpha}_{i,j}^l\right),
\end{equation}
where $\bigcup_{i}\{\boldsymbol{t}_i, w_{\lambda_i}\}$ are the parameters of hypernetworks and the embedding vectors for all the clients that are updated in the central server by minimizing (\ref{pfedla1}). Accordingly, for $C_i$ their update rules can be written as follows:
\begin{equation}
\label{pfedla2}
\begin{aligned}
    \boldsymbol{t}_i &\leftarrow  \boldsymbol{t}_i - \eta(\nabla_{\boldsymbol{t}_i}w_i^*)^T\Delta w_i,\\
    w_{\lambda_i} &\leftarrow  w_{\lambda_i} - \eta(\nabla_{w_{\lambda_i}}w_i^*)^T\Delta w_i.
\end{aligned}
\end{equation}
The reader is referred to \cite{ma2022layer} for the derivation of (\ref{pfedla2}).

\section{A Theoretical Comparison of the Selected\\ Federated Learning Algorithms}
\label{sec:theoretic_comp}

In this section, the selected FL algorithms are theoretically compared based on their: i) local training complexity; ii) aggregation complexity; iii) learning efficiency; iv) communication cost; and v) scalability in terms of number of clients. The comparison of the algorithms under each criterion is given in a separate subsection below, while their overall comparison is summarized in Table~\ref{theoretical_comp}.

\subsection{Local Training Complexity}
The local training complexity of a FL algorithm is associated to the computational cost of training a local DL model per client~\cite{luo2021cost}. Since the FedAvg algorithm employs only the standard empirical risk minimization (ERM) with stochastic gradient descent (SGD) during local FL training, it has the lowest local training complexity. The model aggregation focused FL algorithms (pFedLA, FedNova and FedBN) also utilize the same local training procedure with FedAvg, and thus are associated with the lowest local training complexity. However, the local training focused FL algorithms (FedProx, SCAFFOLD, MOON and FedDC) exploit either proximal terms or gradient adjustment techniques that increases local training complexity compared to other algorithms due to additional calculations of loss functions and gradient terms. In particular, the MOON algorithm is the most computationally complex algorithm since it requires additional forward passes to extract image feature vectors using the global model as well as the local model parameters from the previous FL rounds.

\subsection{Aggregation Complexity}
The aggregation complexity of a FL algorithm refers to the computational cost of aggregating local model parameters from multiple clients to obtain a global model~\cite{so2022lightsecagg}. Since the local training focused FL algorithms exploit the standard iterative model averaging during parameter aggregation as in the FedAvg algorithm, these algorithms are associated with the lowest aggregation complexity. Among the other algorithms, FedNova and FedBN does not significantly increase the computational cost of aggregation as they either normalize or sort out local model parameters during aggregation (which are not costly operations). Thus, they are also associated with the lowest aggregation complexity.
However, the pFedLA algorithm requires to train multiple hypernetworks on the central server for the aggregation stage. This significantly increases the aggregation complexity of pFedLA in proportion to the depth of the considered hypernetworks.

\subsection{Learning Efficiency}
The learning efficiency of a FL algorithm is defined as the minimum number of communication rounds, which is required to achieve a high inference performance of the resulting global model. We would like to remark that the local models do not converge in the early communication rounds. Accordingly, weighting the local model parameters for model parameter aggregation as in the FedNova and pFedLA algorithms reduces the global model performance on the early rounds of FL training. In detail, during the initial rounds using the BN parameters of each local model during the aggregation increases the alignment of the global model with local data characteristics, and thus leads to higher learning efficiency. However, the FedBN algorithm excludes BN parameters during the aggregation. Thus, it reduces the learning efficiency of the global model if the level of training data heterogeneity is not significantly high. Accordingly, the learning efficiency of the local training focused algorithms (FedProx, SCAFFOLD, MOON, FedDC) is expected to be higher than that of model aggregation focused algorithms (FedNova, FedBN, pFedLA). In particular, the learning efficiency of the FedProx algorithm depends on its $\gamma$ hyperparameter. A high value of $\gamma$ leads to minor updates in local training, which reduces the convergence speed. The convergence rate of the MOON algorithm is very similar to the standard iterative model averaging~\cite{li2021model}. However, it performs better in later communication rounds as it is less affected by training data heterogeneity. The learning efficiency of the SCAFFOLD and FedDC algorithms is similar since they both utilize gradient adjustment techniques in the same way. In greater details, FedDC slows down the local convergence in the later communication rounds by adding a proximal term to the local objective functions. We would like to note that the capability of the these algorithms in terms of learning efficiency is highly affected by the training data heterogeneity across clients that may lead the algorithms to show different characteristics under different heterogeneity levels~\cite{luo2022tackling}.
 
We would like to note that all the considered FL algorithms assume that the required computational power for FL training is available. This ensures that the servers of clients participating in FL training are capable of processing input RS data and performing local DL model training. However, using IoT devices or onboard platforms for client servers can lead to longer training times or inability to execute certain FL algorithms since they usually have limited computational power and processing abilities {\cite{vcolakovic2018internet}}. IoT devices and onboard platforms also frequently encounter bandwidth constraints and intermittent connectivity {\cite{misra2020blockchain}}. Such a limited connectivity can result in information loss if DL model updates are delayed, dropped, or corrupted during transmission. The above-mentioned characteristics of IoT devices and onboard platforms may significantly reduce the learning efficiency of FL training independently from the selected FL algorithm. This poses a challenge for FL in RS, which necessitates frequent communication between the clients and the central server.

\subsection{Communication Cost}
The communication cost of a FL algorithm is defined as the amount of parameters required to be transmitted between the considered clients and a central server~\cite{shahid2021communication}. Thus, it increases with the total number of parameters, which include DL model parameters as well as additional parameters (e.g., hyperparameters, gradients etc.) required to be shared with the central server. The FedAvg, FedProx, MOON, pFedLA, FedNova and FedBN algorithms transmit only the DL model parameters between the central server and the clients. Thus, these algorithms are associated with the lowest communication cost. However, the SCAFFOLD and FedDC algorithms require to also transmit the control variate parameters in addition to the model parameters that leads to higher communication cost than the other algorithms. 

It is worth noting that as discussed in the previous subsection, utilizing IoT devices and onboard platforms in the clients generally requires a high number of communication rounds during FL training. This can lead to a significant increase in the communication cost of any FL algorithm in RS. This becomes particularly challenging in the environments with unreliable or slow network connections. To address the constraints posed by limited computational resources and bandwidth, model compression based FL algorithms (e.g., \mbox{\cite{khan2024deep,zhu2023model,xu2022adaptive}}) can be utilized for IoT devices and onboard platforms. The quantized models require less computational power for both training and inference, making them more suitable for deployment on such resource-constrained devices. Moreover, the quantized models result in smaller model updates, requiring less data transmission during communication rounds. This reduces the communication cost. However, while these techniques reduce model size and complexity, they may also lead to inaccurate RS image representation learning due to lossy compression through quantized models.

\subsection{Scalability}
\label{scalability_sec}
The scalability of a FL algorithm is defined as its ability to efficiently handle a large number of participating clients and the corresponding distributed training data while accurately learning a global model~\cite{huba2022papaya}. As the total number of clients increases, there is a higher chance of increase in the variation among the parameters of local models. This may lead to more deviation of the global model from its global optima compared to when there is less number of participating clients to FL. The model aggregation focused FL algorithms do not address the deviation among the local models of clients during the local training stage. Thus, among all the investigated FL algorithms, they are associated with the lowest scalability towards the number of clients. As the FedProx, MOON and FedDC algorithms add proximal terms to local objectives to directly prevent the deviation between the model parameters, their ability to handle a high number of clients is higher than the other algorithms. Since the SCAFFOLD algorithm employs control variate parameters for preventing the client-drift problem that can also limit the deviation of model parameters, it has moderate ability in terms of scalability. 

\section{Data Set Description and Design of Experiments}
\label{sec:exp_setup}
\subsection{Data Set Description}
\label{dataset}
The experiments were conducted on the BigEarthNet-S2 benchmark archive~\cite{sumbul2021bigearthnet}. It includes 590,326 Sentinel-2 images acquired over 10 European countries. For the experiments, we employed a subset of BigEarthNet-S2 that includes images acquired over Austria, Belgium, Finland, Ireland, Lithuania, Serbia and Switzerland. Each image in BigEarthNet-S2 is made up of 120$\times$120 pixels for 10m bands, 60$\times$60 pixels for 20m bands, and 20$\times$20 pixels for 60m bands. 60m bands were not considered in the experiments, while bicubic interpolation was applied to 20m bands. This leads to 120$\times$120 pixels for each of 10 bands. Each image was annotated with multi-labels from the CORINE Land Cover Map (CLC) database of the year 2018. In the experiments, we utilized the 19 class nomenclature presented
in~\cite{sumbul2021bigearthnet}. 

\begin{table}[t]
\renewcommand{\arraystretch}{1.15}
\setlength\tabcolsep{9pt}
\caption{Decentralization Scenarios Designed for the BigEarthNet-S2 Benchmark Archive}
\centering
\begin{tabular}{@{}llc@{}}
\hline
Scenario & \thead[c]{Acquisition Conditions of Each Client}         & \thead[c]{Training Data\\Heterogeneity} \\ \hline
DS1     & Summer season on all countries     & Low          \\
DS2     & Summer season on a specific country   & Medium         \\
DS3     & All seasons on a specific country & High          \\ \hline
\end{tabular}
\label{table:scenarios}
\end{table}
In the experiments, we used the official train and test splits of BigEarthNet-S2. To perform FL on decentralized and unshared training sets, we defined clients by designing three different decentralization scenarios on the official train split of BigEarthNet-S2 as follows: 
\begin{itemize}
    \item \textit{Decentralization Scenario 1 (DS1)}: The images acquired in summer only were randomly distributed to different clients.
    \item \textit{Decentralization Scenario 2 (DS2)}: The images acquired in summer only were distributed in a country specific way (i.e., each client holds the images acquired over only one country).
    \item \textit{Decentralization Scenario 3 (DS3)}: The images acquired in all the seasons were distributed in a country specific way (i.e., each client holds the images acquired over only one country.
\end{itemize}

Due to the design of decentralization scenarios, each scenario is associated with a different level of training data heterogeneity across clients. The lowest level of heterogeneity is present in DS1, because the images in the same season are randomly distributed to clients. The highest level of heterogeneity is present in DS3, because the training data is decentralized based on the countries, while each client can include images acquired on the same area in different seasons. All the scenarios are summarized in Table~\ref{table:scenarios}. 
\input{nb_clients_comp}
\subsection{Design of Experiments}
For the experimental comparison of the selected FL algorithms, we focus our attention on multi-label image classification (MLC) problems in RS. To this end, we employed the ResNet-50 CNN architecture \cite{he2016deep} as the backbone of the global model, which is followed by the fully connected layer of 19 neurons to perform MLC of RS images. For the training of the global model on decentralized and unshared RS image archives by using each selected FL algorithm, we employed the same FL training procedure. Thus, the number $R$ of communication rounds for FL training was set to $40$. For each decentralization scenario, the number $K$ of clients was varied as $K=7,14,28$. For each client, the same DL architecture with the global model was utilized for the corresponding local model. In one communication round, each local model is trained for $E$ epochs with the mini-batch size of $1024$, while $E$ was varied in the range of $[1,7]$ with the step size $2$. For local model training, we used the Adam variant of stochastic gradient descent with the initial learning rate of $10^{-3}$ and the weight decay of $0.9$. All the experiments were performed on NVIDIA A100 80GB GPUs.

We carried out various experiments to: 1) conduct a sensitivity analysis of the selected FL algorithms in terms of the number $E$ of local epochs and the number $K$ of clients; 2) compare the FL algorithms in terms of MLC accuracy, local training complexity and learning efficiency. The results are provided in terms of $F_1$ score on the official test split of BigEarthNet-S2 for MLC of RS images. The local training complexity of each algorithm is evaluated in terms of its computation time in seconds to complete one communication round of local training. For all the selected FL algorithms, we used the same DL architecture for a fair comparison. For the hyperparameter selection of each algorithm, we followed the same procedure suggested in the corresponding paper. To this end, the hyperparameters $\gamma$ of the FedProx algorithm and $\tau$ of the MOON algorithm were set to $0.01$ and $1$, respectively. The weight of the model-contrastive loss function $\mathcal{L}_{\text{MC}}$ in the MOON algorithm was set to $0.1$.

\section{Experimental Results}
\label{sec:exp_comp}
We performed experiments in order to:
1) conduct a sensitivity analysis; and 2) compare the effectiveness of the investigated FL algorithms (FedAvg \cite{mcmahan2017communication}, FedProx \cite{li2020federated}, SCAFFOLD \cite{karimireddy2020scaffold}, MOON \cite{li2021model}, FedDC \cite{gao2022feddc}, FedNova \cite{wang2021novel}, pFedLA \cite{ma2022layer}, and FedBN \cite{li2021fedbn}). 

\subsection{Sensitivity Analysis of the Selected FL Algorithms}
In the first set of trials, we analyzed the effectiveness of the FL algorithms in terms of the number $E$ of local epochs. Fig. \ref{fig:sens} shows the $F_1$ scores under the decentralization scenario 1 when different numbers $E$ of local epochs are used in the FedProx algorithm. Once can observe from the figure that when $E=3$, the FedProx algorithm achieves the highest $F_1$ scores in the latest rounds of the FL training. From the figure, one can also see that when each local training data is fed to the corresponding local model only once within each communication round (i.e., $E=1$), the FedProx algorithm obtains the lowest $F_1$ scores compared to the higher numbers of local epochs. This is due to the fact that when $E=1$, the change in the local model parameter updates decreases throughout communication rounds compared to higher numbers of epochs. This leads to a slow convergence of the global model. When $E>3$ (i.e., $E=5$ or $E=7$), the divergence among local model parameters increases. This leads to deviation of the global model from its global optima after parameter averaging, and thus ineffective MLC performance. We would like to note that the results of the sensitivity analysis for the number of local epochs were also confirmed through the experiments for the FedAvg, SCAFFOLD, MOON, FedDC, pFedLA, FedNova and FedBN algorithms under all the scenarios (not reported for space constraints). Thus, we set $E$ to $3$ for the rest of experiments.
\input{sensitivity_analysis.tex}

In the second set of trials, we assessed the effectiveness of the FL algorithms in terms of the number $K$ of clients. Table \ref{tab:DS_33} shows the $F_1$ scores of FL algorithms when different numbers $K$ of clients are considered under all the scenarios. One can observe from the table that as $K$ increases, most of the algorithms lead to slightly lower $F_1$ scores. This is due to the increase in the variation of the local model parameters when $K$ increases. The effect of the number of clients on the performances of the algorithms is more evident when the number of clients is increased to $28$ from $14$ (i.e., $K=14\rightarrow K=28$). As an example, under DS1, the $F_1$ score of the FedNova algorithm decreases $0.2\%$ when $K=7\rightarrow K=14$, while it decreases $1\%$ when $K=14\rightarrow K=28$. This is due to the fact that the variation among the local models increases more than the increase in the number of clients. In greater details, when the training data heterogeneity across clients increases (i.e., DS1$\rightarrow$DS3), the ability of the algorithms to overcome the higher number of clients (i.e., scalability) generally decreases. As an example, when $K=7\rightarrow K=28$, the $F_1$ score of the MOON algorithm decreases $1.6\%$ and $4.2\%$ under DS1 and DS2, respectively. This is due to fact that as the number of clients increases when training data is distributed across more heterogeneous clients, the variation among the local models further increases. When the FL algorithms are compared among each other in terms of their scalability towards the number of clients, the decrease on $F_1$ scores obtained by the local training focused algorithms are generally lower than the other algorithms. However, the ability of these algorithms towards $K$ is highly affected by the training data heterogeneity across clients, as it can be seen from the table. This is inline with our theoretical analysis in the Section \ref{scalability_sec}. We would like to note that for the rest of experiments, we set the number $K$ of clients as $K=7$ for the sake of simplicity. 
\begin{table}[t]
\renewcommand{\arraystretch}{1.2}
\setlength\tabcolsep{13pt}
\caption{$F_1$ Scores (\%) Obtained by the FL Algorithms Under Decentralization Scenarios}
\label{tab:DS_1}
\centering
\begin{tabular}{@{}llccc@{}}
\hline
Algorithm & & DS1 & DS2 & DS3 \\ \hline
FedAvg \cite{mcmahan2017communication} & & {79.2}    & {52.2}    & {47.5} \\ \hline
\multirow{4}{*}{\thead[l]{Local\\Training\\Focused}} & FedProx \cite{li2020federated} & {79.5}    & {56.3}  & {53.3} \\ 
& SCAFFOLD \cite{karimireddy2020scaffold}& {78.6}    & {57.2}    & {54.4}  \\ 
& MOON \cite{li2021model}  & {\textbf{81.7}}    & {59.5}    & {55.2}\\ 
& FedDC \cite{gao2022feddc}  & {78.8}    & {58.4}    & {\textbf{57.4}}\\ \hline
\multirow{3}{*}{\thead[l]{Model\\Aggregation\\Focused}}  & FedNova \cite{wang2021novel} & {80.6}    & {53.7}  & {48.6}\\ 
& pFedLA \cite{ma2022layer} & {81.3}    & {56.8}    & {53.6}\\ 
 & FedBN \cite{li2021fedbn}  & {78.4}    & {\textbf{62.3}}    & {57.1}\\ \hline
\end{tabular}
\end{table}
\subsection{Comparison of the Selected FL Algorithms}

In the first set of trials, we compared the effectiveness of the selected FL algorithms in the context of MLC of RS images. Table \ref{tab:DS_1} shows the corresponding $F_1$ scores under decentralization scenarios 1 (DS1), 2 (DS2) and 3 (DS3). One can assess from the table that as training data heterogeneity across clients increases through our decentralization scenarios, each algorithm leads to lower $F_1$ scores. As an example, the FedNova algorithm achieves more than $25\%$ and $30\%$ higher scores under DS1 compared to those under DS2 and DS3, respectively. In greater details, under DS1, the $F_1$ scores obtained by the algorithms are close to each other. This is due to the low training data heterogeneity level in DS1, for which the considered FL algorithms do not significantly increase the MLC performance. The algorithms that add proximal terms to the local objective function (i.e., FedProx and MOON) achieve slightly higher results than those obtained by the algorithms based on gradient adjustment techniques (i.e., SCAFFOLD and FedDC) under DS1. As an example, the FedProx algorithm outperforms the FedDC algorithm by 0.4\% higher $F_1$ score under DS1. In particular, the highest $F_1$ score was obtained by the MOON algorithm, which is $3.3\%$ higher than that obtained by the worst performing algorithm (i.e., FedBN) under DS1. This indicates that the elimination of batch normalization parameters from model parameter aggregation at the central server decreases MLC performance when the level of data heterogeneity is not significantly high. However, as the training data heterogeneity increases, the differences in $F_1$ scores obtained by the algorithms significantly increases. As the worst performing algorithm under DS1, the FedBN algorithm achieves the highest $F_1$ score under DS2. This shows that FedBN provides more accurate MLC results compared the other algorithms when the concept shift is high. We would like to note that the FedAvg algorithm leads to the lowest scores under both DS2 and DS3 as its learning process is highly interfered by non-IID training data. While comparing the algorithms under DS3, the FedDC algorithm achieves the highest $F_1$ score of $57.4\%$, which is only $1\%$ lower than its performance under DS2. This indicates that the FedDC algorithm is more robust to training data heterogeneity across clients than the other algorithms. 

\begin{table}[t]
\renewcommand{\arraystretch}{1.1}
\setlength\tabcolsep{17pt}
\caption{Computation Times (in seconds) Required by the FL Algorithms to Complete One Communication Round of Local Training}
\label{time}
\centering
\begin{tabular}{@{}llc@{}}
\hline
Algorithm & & \thead[c]{Computation Time\\(Local Training)} \\\hline
FedAvg~\cite{mcmahan2017communication} & & \textbf{184} \\\hline
\multirow{4}{*}{\thead[l]{Local\\Training\\Focused}} & FedProx~\cite{li2020federated} & 189 \\
& SCAFFOLD~\cite{karimireddy2020scaffold} & 199\\
& MOON~\cite{li2021model} & 258\\ 
& FedDC~\cite{gao2022feddc} & 199\\\hline
\multirow{3}{*}{\thead[l]{Model\\Aggregation\\Focused}} & 
FedNova~\cite{wang2021novel} & \multirow{3}{*}{\textbf{184}}\\ 
& pFedLA~\cite{ma2022layer} & \\ 
& FedBN~\cite{li2021fedbn} & \\ \hline
\end{tabular}
\end{table}
In the second set of trials, we compared the effectiveness of the considered FL algorithms in terms of their local training complexity. Table \ref{time} shows the computation times (in seconds) required by all the algorithms to complete one communication round of local training. As one can observe from the table that the computation time required by the FedAvg, FedNova, FedBN and pFedLA algorithms is the same and equal to $184$ seconds, which is lower than those required by other algorithms. This is due to the fact that FedNova, FedBN and pFedLA are model aggregation focused algorithms and apply the same local training strategy with the FedAvg algorithm. The computation time required by the FedProx algorithm is $189$ seconds, which is slightly higher than those required by the model aggregation focused algorithms. This is due to calculating the $L_2$ distance between global and local models in FedProx. The computation time required by the SCAFFOLD and FedDC algorithms is slightly increased compared to that by the FedProx algorithm due to the calculation of the control variate parameters. Since the MOON algorithm extracts the feature vectors of an RS image using three different models, its local training complexity is the highest among other algorithms with the computation time of $258$ seconds. 

We would like to note that overall FL training time is also affected by the number $R$ of communication rounds in addition to the computation time per round. The FL algorithms, which are capable of efficiently learning global models, can achieve a high MLC performance under a small number of communication rounds. In the third set of trials, we compared the effectiveness of the FL algorithms in terms of learning efficiency. Fig \ref{fig:le} shows the $F_1$ scores of the algorithms for the first ten communication rounds under DS1. One can observe from the figure that the local training focused algorithms (FedProx, SCAFFOLD, MOON and FedDC) achieve higher $F_1$ scores at earlier communication rounds compared the model aggregation focused algorithms. As an example, the MOON reaches almost $70\%$ $F_1$ score at the 3rd communication round, while FedNova, FedBN and pFedLA require 5, 10 and 6 communication rounds, respectively, to achieve the same performance. In greater details, after 4 communication rounds the performance of the FedAvg algorithm reaches to that of MOON. This shows that the learning efficiency of FedAvg is also high. Although the selected algorithms that restrict local training by adding proximal terms to the objective function (i.e., FedProx, MOON, FedDC) alleviate the effects of training data heterogeneity, these algorithms slow down the local convergence and limit learning efficiency. Since DS1 is associated with the lowest training data heterogeneity compared to other scenarios, the FedAvg algorithm achieves higher scores in the initial communication rounds than these algorithms. Among the algorithms that utilize gradient adjustment techniques during local training SCAFFOLD shows the highest learning efficiency. This proves that the gradient adjustment using control variates speeds up the convergence.

\input{learning_efficiency.tex}
\section{Conclusion and Discussion}
\label{sec:conclusion}
In this paper, as a first time in RS, we have presented a comparative study of advanced FL algorithms. To this end, we initially provided the systematic review of the FL algorithms, which are presented in the computer vision community. In accordance with this review, we have selected the state-of-the-art FL algorithms FedProx \cite{li2020federated}, SCAFFOLD \cite{karimireddy2020scaffold}, MOON \cite{li2021model}, FedDC \cite{gao2022feddc}, FedNova \cite{wang2021novel}, pFedLA \cite{ma2022layer}, and FedBN \cite{li2021fedbn} based on their effectiveness with respect to training data heterogeneity accross clients. We have provided the extensive overview of the selected algorithms in the framework of RS image classification, while analyzing their capability to address the challenges of non-IID training data. The considered algorithms have been theoretically compared in terms of their: 1) local training complexity; 2) aggregation complexity; 3) learning efficiency; 4) communication cost; and 5) scalability in terms of number of clients. In addition to the theoretical analyses of these algorithms, we have also experimentally compared them under different decentralization scenarios. For our experimental analyses, we have focused our attention on multi-label classification problems. We would like to note that all the selected FL algorithms have been considered for the first time in RS. Based on our theoretical and experimental comparison among these algorithms, we have derived a guideline for selecting suitable algorithms in RS as follows:
\begin{enumerate}
 \item The algorithms achieve similar performances when the level of training data heterogeneity is low. Thus, the algorithms associated with lower local training complexity (FedAvg, FedNova, FedBN and pFedLA) can be selected if the training data is not highly non-IID among clients.
 \item As the training data heterogeneity increases, there is a significant drop in the performance of all the algorithms. However, the MOON, FedDC and FedBN algorithms are more effective towards non-IID training data than the other algorithms. Thus, these algorithms can be selected when the training data is highly non-IID among clients. 
 \item Although MOON performs better than most of the algorithms, its training complexity is the highest among the considered algorithms. Thus, as an alternative, FedBN and FedDC can be selected if the local training complexity needs to be limited. 
 \item The aggregation focused algorithms provide comparable results with the local training focused algorithms. Although their local training complexity is lower, their aggregation complexity is higher than the local training focused algorithms. In particular, the pFedLA algorithm significantly increases the aggregation complexity when the number of clients is high. Thus, FedBN and FedNova can be be selected among the aggregation focused algorithms when the aggregation complexity needs to be limited.
\end{enumerate}

It is worth noting that we have compared these algorithms also in the context of single-label image scene classification problems using EuroSAT benchmark archive {\cite{eurosat}}. From the experimental results, we have observed the same relative behavior in the results obtained using the BigEarthNet-S2 archive in the context of multi-label classification problems. Thus, in order not to increase the complexity of the manuscript, the results on the EuroSAT benchmark archive is not reported. We would like to also note that the above-mentioned algorithms can also be easily adapted for other RS image analysis tasks (e.g., change detection, biophysical parameter estimation, etc.). This can be achieved by using the corresponding loss functions and the task heads in the considered DL model. 

It is worth emphasizing that the considered FL algorithms embody a significant potential for knowledge discovery from distributed RS image archives. We believe that in the upcoming years FL from decentralized RS image archives will attract more attention that will give rise to new challenges and new research directions. The integration of multi-modal learning into FL algorithms for RS applications is an important future research direction to effectively leverage unshared multi-modal RS data distributed across clients {\cite{Buyuktas:2023}}. This will pose another level of training data heterogeneity resulting from different RS data modalities. As an example of multi-modal RS data, RS images acquired by different sensors with distinct characteristics (e.g., passive multispectral and hyperspectral sensors, and Synthetic Aperture Radar (SAR) active instruments) can embody profound differences, and thus significantly increase the training data heterogeneity when such images are distributed across clients. Personalized and adaptive FL model training is a crucial area for future research in RS. This is due to the fact that for many RS applications the clients participating to FL training can be highly diverse in terms of privacy requirements and computational powers (e.g., regional agencies versus individual farmers in the context of crop monitoring). This research direction includes the development of FL algorithms for RS tailored to: i) client-specific model customization; ii) dynamic model updating that takes into account the individual requirements of each client; and iii) transfer learning techniques that facilitate knowledge transfer among distributed clients. We think that the development of FL systems will be crucial also in enhancing communication within satellite networks. This can leverage the local computational power of satellites (especially in low Earth orbit constellations), and thus eliminate the need for data transmission to ground stations {\cite{Wu:2024}}. The challenges of applying FL in satellite networks embody not only those of IoT devices, which are mainly limited computational power and processing abilities, bandwidth constraints and intermittent connectivity (see Section IV for details) with probably higher intensity but also dynamics of satellite communication.
   
\section*{Acknowledgment}
This work is supported by the European Research Council (ERC) through the ERC-2017-STG BigEarth Project under Grant 759764.
\bibliographystyle{IEEEtran}
\bibliography{papers.bib}

\begin{IEEEbiography}[{\includegraphics[width=1in,height=1.25in,clip,keepaspectratio]{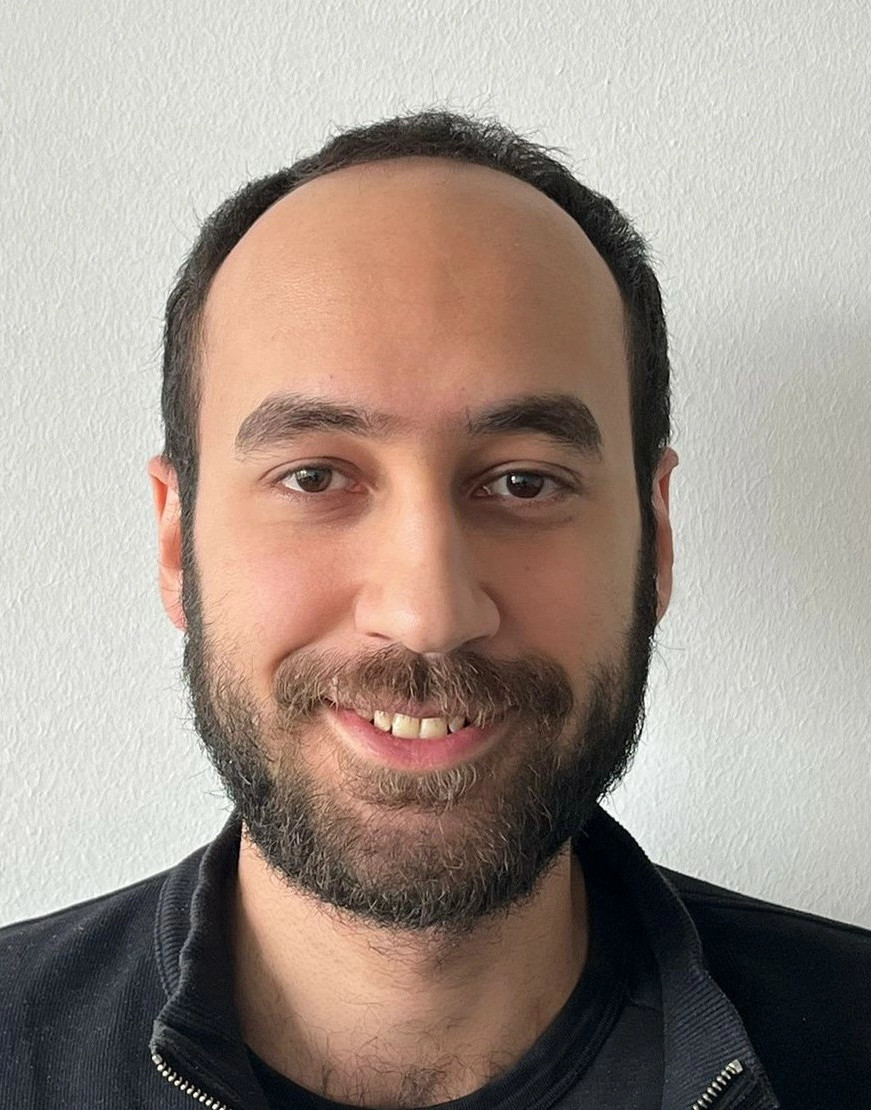}}]{Barı\c{s} B\"{u}y\"{u}kta\c{s}} received the B.S. and M.S. degrees in electrical and electronics engineering from Ozyegin University, Istanbul, Turkey in 2018 and 2020, respectively. He is currently pursuing a Ph.D. degree at the Faculty of Electrical Engineering and Computer Science, Technische Universit\"at Berlin, Germany. He is currently a research associate in the Remote Sensing Image Analysis (RSiM) group, Technische Universit\"at Berlin. His research interests include federated learning, computer vision, pattern recognition, machine learning, and remote sensing. 
\end{IEEEbiography} 

\begin{IEEEbiography}[{\includegraphics[width=1in,height=1.25in,clip,keepaspectratio]{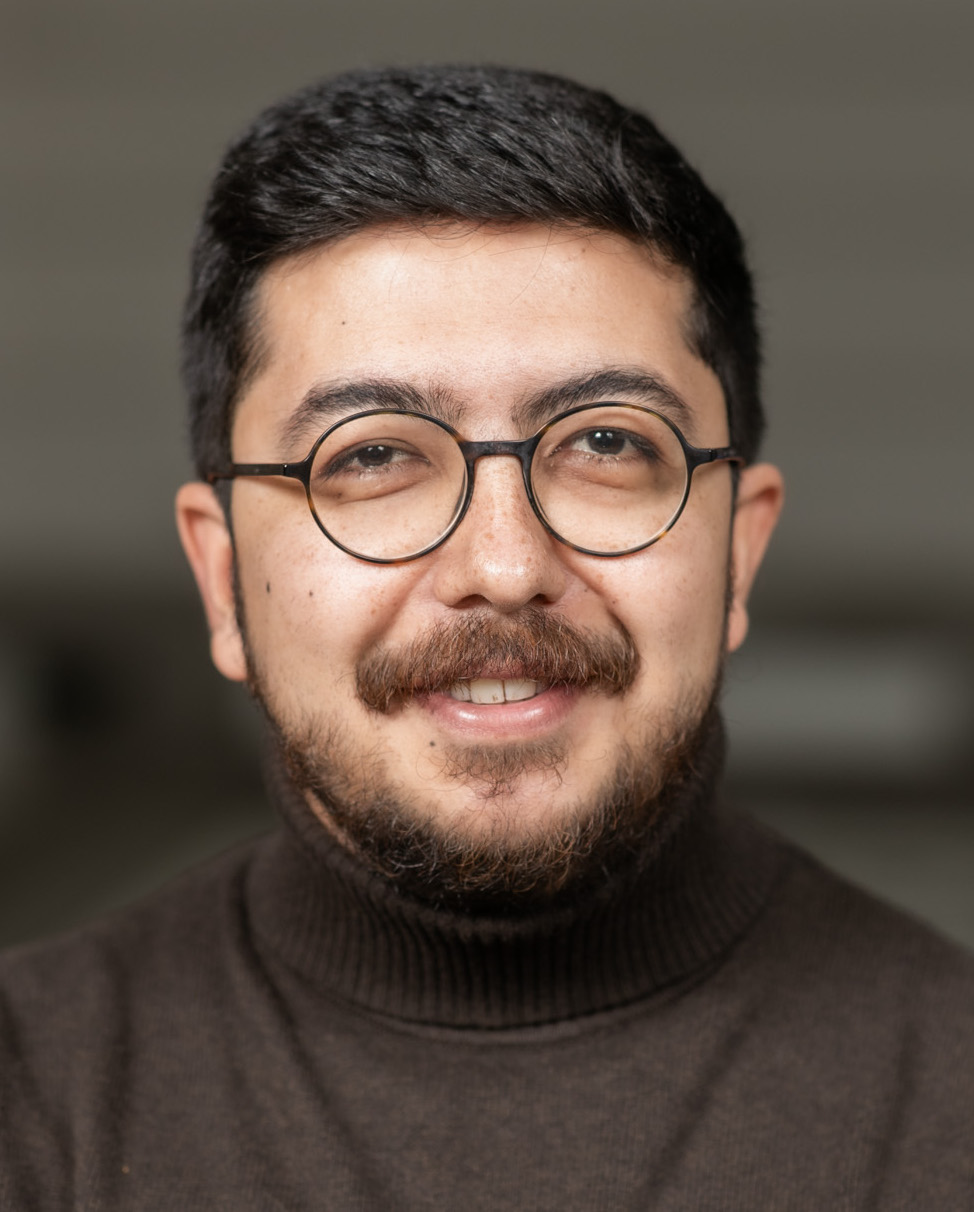}}]{Gencer Sumbul} received his B.S. degree in computer engineering from Bilkent University, Ankara, Turkey in 2015, the M.S. degree in computer engineering from Bilkent University in 2018, and the Ph.D. degree from the Faculty of Electrical Engineering and Computer Science, Technische Universit\"at Berlin, Germany in 2023. He is currently a postdoctoral scientist in the Environmental Computational Science and Earth Observation Laboratory (ECEO), École Polytechnique Fédérale de Lausanne (EPFL). His research interests include computer vision, pattern recognition and machine learning, with special interest in deep learning, large-scale image understanding and remote sensing. He is a referee for journals such as the IEEE Transactions on Image Processing, IEEE Transactions on Geoscience and Remote Sensing, the IEEE Access, the IEEE Geoscience and Remote Sensing Letters, the ISPRS Journal of Photogrammetry and Remote Sensing and international conferences such as European Conference on Computer Vision and IEEE International Geoscience and Remote Sensing Symposium.
\end{IEEEbiography} 
\begin{IEEEbiography}[{\includegraphics[width=1in,height=1.25in,clip,keepaspectratio]{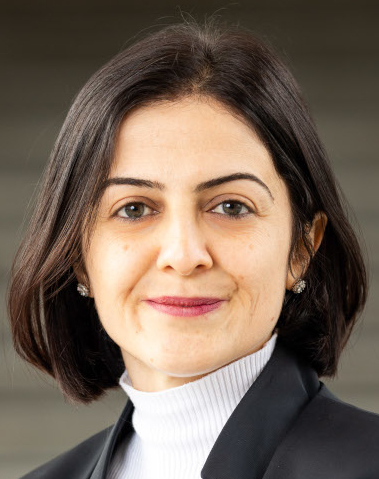}}]{Beg\"{u}m Demir} (S'06-M'11-SM'16) received the B.S., M.Sc., and Ph.D. degrees in electronic and telecommunication engineering from Kocaeli University, Kocaeli, Turkey, in 2005, 2007, and 2010, respectively.

She is currently a Full Professor and the founder head of the Remote Sensing Image Analysis (RSiM) group at the Faculty of Electrical Engineering and Computer Science, TU Berlin and the head of the Big Data Analytics for Earth Observation research group at the Berlin Institute for the Foundations of Learning and Data (BIFOLD). Her research activities lie at the intersection of machine learning, remote sensing and signal processing. Specifically, she performs research in the field of processing and analysis of large-scale Earth observation data acquired by airborne and satellite-borne systems. She was awarded by the prestigious ‘2018 Early Career Award’ by the IEEE Geoscience and Remote Sensing Society for her research contributions in machine learning for information retrieval in remote sensing. In 2018, she received a Starting Grant from the European Research Council (ERC) for her project “BigEarth: Accurate and Scalable Processing of Big Data in Earth Observation”. She is an IEEE Senior Member and Fellow of European Lab for Learning and Intelligent Systems (ELLIS).

Dr. Demir is a Scientific Committee member of several international conferences and workshops, such as: Conference on Content-Based Multimedia Indexing, Conference on Big Data from Space, Living Planet Symposium, International Joint Urban Remote Sensing Event, SPIE International Conference on Signal and Image Processing for Remote Sensing, Machine Learning for Earth Observation Workshop organized within the ECML/PKDD. She is a referee for several journals such as the PROCEEDINGS OF THE IEEE, the IEEE TRANSACTIONS ON GEOSCIENCE AND REMOTE SENSING, the IEEE GEOSCIENCE AND REMOTE SENSING LETTERS, the IEEE TRANSACTIONS ON IMAGE PROCESSING, Pattern Recognition, the IEEE TRANSACTIONS ON CIRCUITS AND SYSTEMS FOR VIDEO TECHNOLOGY, the IEEE JOURNAL OF SELECTED TOPICS IN SIGNAL PROCESSING, the International Journal of Remote Sensing), and several international conferences. Currently she is an Associate Editor for the IEEE GEOSCIENCE AND REMOTE SENSING LETTERS, MDPI Remote Sensing and International Journal of Remote Sensing.
\end{IEEEbiography}
\vfill
\end{document}

%% file: sources.tex
\usepackage[linesnumbered,lined,boxed,commentsnumbered,ruled]{algorithm2e}
\SetKwComment{Comment}{$\triangleright$ }{}
\SetKwInOut{Input}{Input}\SetKwInOut{Output}{Output}
\usepackage{etoolbox}

\SetCommentSty{mycommfont}

\makeatletter
\patchcmd{\@algocf@start}
  {-1.5em}
  {0pt}
  {}{}
\makeatother

\usepackage{amsmath,amssymb,amsfonts}
\usepackage[usenames,dvipsnames]{xcolor} 
\usepackage{multirow}
\usepackage{float}
\usepackage{pdfpages}
\usepackage{amsmath}
\usepackage{comment}
\usepackage{float}
\usepackage{bbm}
\usepackage{multicol}
\usepackage{adjustbox}
\usepackage{amssymb}
\usepackage{amsfonts}
\usepackage{multicol}
\usepackage{pifont}
\usepackage{makecell}
\def\cmark{\ding{51}} 
\def\xmark{\ding{55}}
\usepackage{mathtools}
\usepackage{upgreek}
\usepackage[noadjust]{cite}
\usepackage{url}
\usepackage{pgfplots}
\usepgfplotslibrary{colorbrewer}
\pgfplotsset{cycle list/Set1-9}
\tikzset{every picture/.style={line width=1pt}}
\usetikzlibrary{patterns}
\usepgfplotslibrary{fillbetween}
\usepackage{color,soul}

\pgfplotsset{compat=1.18}
\pgfplotsset{
tick label style = {font=\sffamily\scriptsize},
every axis label = {font=\sffamily\scriptsize},
legend style = {font=\sffamily\scriptsize},
}
\usepackage{tikz}

\colorlet{green_fl}{green!40!black}
\colorlet{yellow_fl}{Yellow!50!Brown}
\colorlet{red_fl}{red!50!black}

\colorlet{FedAvg_c}{Gray}
\colorlet{FedProx_c}{Brown}
\colorlet{SCAFFOLD_c}{Fuchsia}
\colorlet{MOON_c}{Cyan!70!black}
\colorlet{FedDC_c}{orange}
\colorlet{pFedLA_c}{BlueViolet}
\colorlet{FedNova_c}{pink!60!red}
\colorlet{FedBN_c}{Magenta}

\definecolor{fedAvg1}{RGB}{53,88,13}
\definecolor{fedAvg2}{RGB}{134,24,24}
\definecolor{fedAvg3}{RGB}{132,28,153}
\definecolor{fedAvg4}{RGB}{28,74,128}

\newcommand*\numcircled[1]{\raisebox{.5pt}{\textcircled{\raisebox{-.9pt} {#1}}}}

\ifCLASSINFOpdf

\else

\fi

%% file: FL_figure.tex
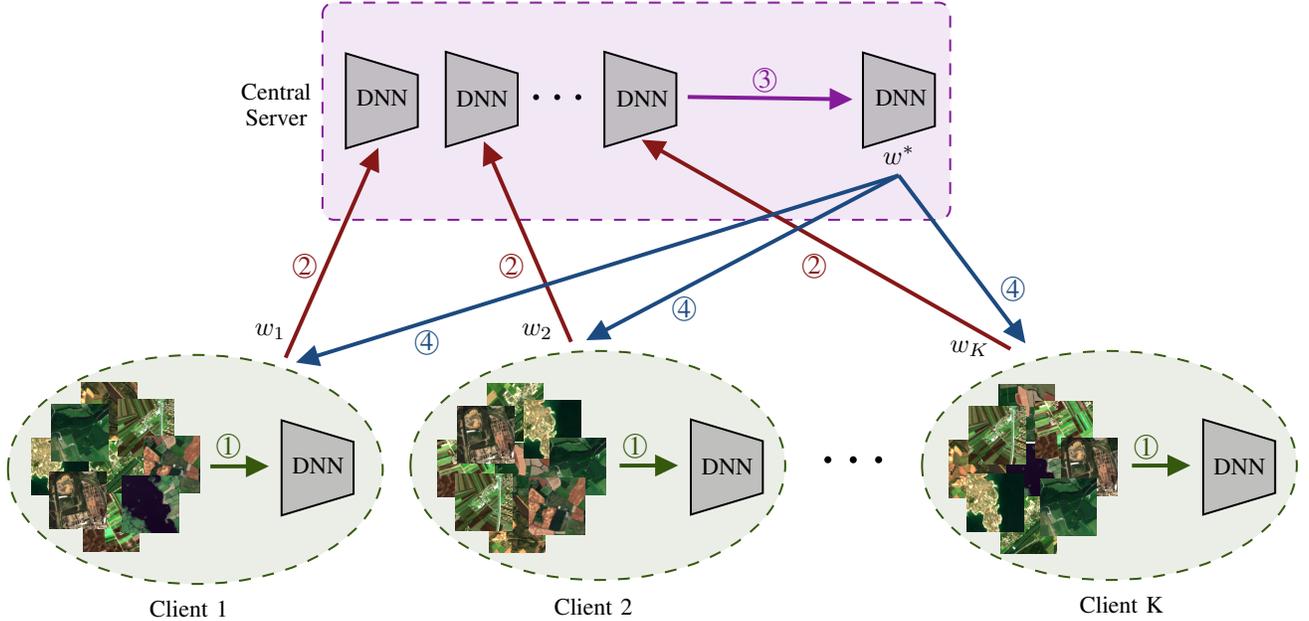
\begin{figure*}[t]
    \centering

\input{FL_figure_source}

    \caption{An illustration of federated learning applied with the iterative model averaging strategy to learn a deep neural network (DNN) when there is no access to training data on decentralized RS image archives (i.e., clients). There are four main steps of iterative model averaging: \textcolor{fedAvg1}{\numcircled{1}} each client updates its DNN with the corresponding local data, leading to one set $w_k$ of local DNN parameters for the $k$th client; \textcolor{fedAvg2}{\numcircled{2}} each client sends its local DNN parameters to the central server; \textcolor{fedAvg3}{\numcircled{3}} the central server aggregates the parameters of all local models, aiming to obtain the set $w^*$ of optimum parameters of the global DNN; and \textcolor{fedAvg4}{\numcircled{4}} the central server sends the parameters of the global DNN to each client. These steps are repeated several times until the convergence of the global DNN.}
\label{figure}
\end{figure*}

%% file: FL_figure_source.tex
\tikzset{every picture/.style={line width=0.75pt}} 

\begin{tikzpicture}[x=0.75pt,y=0.75pt,yscale=-1,xscale=1]

\draw  [color={rgb, 255:red, 53; green, 88; blue, 13 }  ,draw opacity=1 ][fill={rgb, 255:red, 53; green, 88; blue, 13 }  ,fill opacity=0.1 ][dash pattern={on 4.5pt off 4.5pt}] (7,238.2) .. controls (7,206.28) and (49.31,180.4) .. (101.5,180.4) .. controls (153.69,180.4) and (196,206.28) .. (196,238.2) .. controls (196,270.12) and (153.69,296) .. (101.5,296) .. controls (49.31,296) and (7,270.12) .. (7,238.2) -- cycle ;
\draw  [color={rgb, 255:red, 53; green, 88; blue, 13 }  ,draw opacity=1 ][fill={rgb, 255:red, 53; green, 88; blue, 13 }  ,fill opacity=0.1 ][dash pattern={on 4.5pt off 4.5pt}] (210,236.2) .. controls (210,204.28) and (252.31,178.4) .. (304.5,178.4) .. controls (356.69,178.4) and (399,204.28) .. (399,236.2) .. controls (399,268.12) and (356.69,294) .. (304.5,294) .. controls (252.31,294) and (210,268.12) .. (210,236.2) -- cycle ;
\draw  [color={rgb, 255:red, 53; green, 88; blue, 13 }  ,draw opacity=1 ][fill={rgb, 255:red, 53; green, 88; blue, 13 }  ,fill opacity=0.1 ][dash pattern={on 4.5pt off 4.5pt}] (468,236.2) .. controls (468,204.28) and (510.31,178.4) .. (562.5,178.4) .. controls (614.69,178.4) and (657,204.28) .. (657,236.2) .. controls (657,268.12) and (614.69,294) .. (562.5,294) .. controls (510.31,294) and (468,268.12) .. (468,236.2) -- cycle ;
\draw  [color={rgb, 255:red, 132; green, 28; blue, 153 }  ,draw opacity=1 ][fill={rgb, 255:red, 132; green, 28; blue, 153 }  ,fill opacity=0.1 ][dash pattern={on 4.5pt off 4.5pt}] (165.63,8.56) .. controls (165.63,5.27) and (168.3,2.6) .. (171.59,2.6) -- (476.24,2.6) .. controls (479.53,2.6) and (482.2,5.27) .. (482.2,8.56) -- (482.2,106.24) .. controls (482.2,109.53) and (479.53,112.2) .. (476.24,112.2) -- (171.59,112.2) .. controls (168.3,112.2) and (165.63,109.53) .. (165.63,106.24) -- cycle ;
\draw  [fill={rgb, 255:red, 155; green, 155; blue, 155 }  ,fill opacity=0.55 ] (145,213.4) -- (181.3,224.29) -- (181.3,250.51) -- (145,261.4) -- cycle ;
\draw [color={rgb, 255:red, 53; green, 88; blue, 13 }  ,draw opacity=1 ][line width=1.5]    (109.07,236.67) -- (134.6,236.61) ;
\draw [shift={(138.6,236.6)}, rotate = 179.87] [fill={rgb, 255:red, 53; green, 88; blue, 13 }  ,fill opacity=1 ][line width=0.08]  [draw opacity=0] (11.61,-5.58) -- (0,0) -- (11.61,5.58) -- cycle    ;
\draw [color={rgb, 255:red, 134; green, 24; blue, 24 }  ,draw opacity=1 ][line width=1.5]    (147,181.8) -- (192.18,79.46) ;
\draw [shift={(193.8,75.8)}, rotate = 113.82] [fill={rgb, 255:red, 134; green, 24; blue, 24 }  ,fill opacity=1 ][line width=0.08]  [draw opacity=0] (11.61,-5.58) -- (0,0) -- (11.61,5.58) -- cycle    ;
\draw [color={rgb, 255:red, 134; green, 24; blue, 24 }  ,draw opacity=1 ][line width=1.5]    (291,173.8) -- (248.59,76.27) ;
\draw [shift={(247,72.6)}, rotate = 66.5] [fill={rgb, 255:red, 134; green, 24; blue, 24 }  ,fill opacity=1 ][line width=0.08]  [draw opacity=0] (11.61,-5.58) -- (0,0) -- (11.61,5.58) -- cycle    ;
\draw [color={rgb, 255:red, 134; green, 24; blue, 24 }  ,draw opacity=1 ][line width=1.5]    (330.08,73.77) -- (512.6,177.4) ;
\draw [shift={(326.6,71.8)}, rotate = 29.59] [fill={rgb, 255:red, 134; green, 24; blue, 24 }  ,fill opacity=1 ][line width=0.08]  [draw opacity=0] (11.61,-5.58) -- (0,0) -- (11.61,5.58) -- cycle    ;
\draw [color={rgb, 255:red, 28; green, 74; blue, 128 }  ,draw opacity=1 ][line width=1.5]    (156.02,183.01) -- (456.2,89.8) ;
\draw [shift={(152.2,184.2)}, rotate = 342.75] [fill={rgb, 255:red, 28; green, 74; blue, 128 }  ,fill opacity=1 ][line width=0.08]  [draw opacity=0] (11.61,-5.58) -- (0,0) -- (11.61,5.58) -- cycle    ;
\draw [color={rgb, 255:red, 28; green, 74; blue, 128 }  ,draw opacity=1 ][line width=1.5]    (302.14,171.13) -- (456.2,89.8) ;
\draw [shift={(298.6,173)}, rotate = 332.17] [fill={rgb, 255:red, 28; green, 74; blue, 128 }  ,fill opacity=1 ][line width=0.08]  [draw opacity=0] (11.61,-5.58) -- (0,0) -- (11.61,5.58) -- cycle    ;
\draw [color={rgb, 255:red, 28; green, 74; blue, 128 }  ,draw opacity=1 ][line width=1.5]    (456.2,89.8) -- (517.79,171.41) ;
\draw [shift={(520.2,174.6)}, rotate = 232.96] [fill={rgb, 255:red, 28; green, 74; blue, 128 }  ,fill opacity=1 ][line width=0.08]  [draw opacity=0] (11.61,-5.58) -- (0,0) -- (11.61,5.58) -- cycle    ;
\draw (62.15,238.51) node  {\includegraphics[width=21.21pt,height=21.54pt]{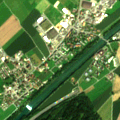}};
\draw (57.99,208.54) node  {\includegraphics[width=21.21pt,height=21.54pt]{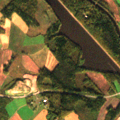}};
\draw (59.1,265.04) node  {\includegraphics[width=21.21pt,height=21.54pt]{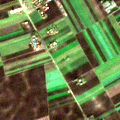}};
\draw (33.14,236.77) node  {\includegraphics[width=21.21pt,height=21.54pt]{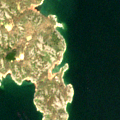}};
\draw (76.77,217.07) node  {\includegraphics[width=21.21pt,height=21.54pt]{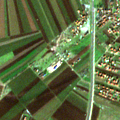}};
\draw (42.19,253.84) node  {\includegraphics[width=21.21pt,height=21.54pt]{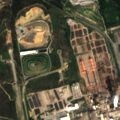}};
\draw (43.16,219.7) node  {\includegraphics[width=21.21pt,height=21.54pt]{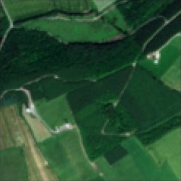}};
\draw (89.66,235.79) node  {\includegraphics[width=21.21pt,height=21.54pt]{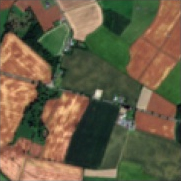}};
\draw (78.95,255.91) node  {\includegraphics[width=21.21pt,height=21.54pt]{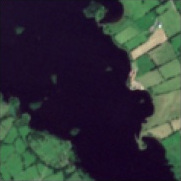}};
\draw (266.95,239.31) node  {\includegraphics[width=21.21pt,height=21.54pt]{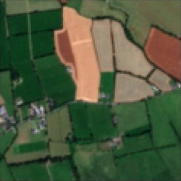}};
\draw (262.79,209.34) node  {\includegraphics[width=21.21pt,height=21.54pt]{figures/1.png}};
\draw (263.9,265.84) node  {\includegraphics[width=21.21pt,height=21.54pt]{figures/2.png}};
\draw (237.94,237.57) node  {\includegraphics[width=21.21pt,height=21.54pt]{figures/3.png}};
\draw (281.57,217.87) node  {\includegraphics[width=21.21pt,height=21.54pt]{figures/4.png}};
\draw (246.99,254.64) node  {\includegraphics[width=21.21pt,height=21.54pt]{figures/5.png}};
\draw (247.96,220.5) node  {\includegraphics[width=21.21pt,height=21.54pt]{figures/6.png}};
\draw (294.46,236.59) node  {\includegraphics[width=21.21pt,height=21.54pt]{figures/7.jpg}};
\draw (283.75,256.71) node  {\includegraphics[width=21.21pt,height=21.54pt]{figures/8.jpg}};
\draw (524.95,239.71) node  {\includegraphics[width=21.21pt,height=21.54pt]{figures/9.jpg}};
\draw (520.79,209.74) node  {\includegraphics[width=21.21pt,height=21.54pt]{figures/10.jpg}};
\draw (521.9,266.24) node  {\includegraphics[width=21.21pt,height=21.54pt]{figures/1.png}};
\draw (495.94,237.97) node  {\includegraphics[width=21.21pt,height=21.54pt]{figures/2.png}};
\draw (539.57,218.27) node  {\includegraphics[width=21.21pt,height=21.54pt]{figures/3.png}};
\draw (504.99,255.04) node  {\includegraphics[width=21.21pt,height=21.54pt]{figures/4.png}};
\draw (505.96,220.9) node  {\includegraphics[width=21.21pt,height=21.54pt]{figures/5.png}};
\draw (552.46,236.99) node  {\includegraphics[width=21.21pt,height=21.54pt]{figures/6.png}};
\draw (541.75,257.11) node  {\includegraphics[width=21.21pt,height=21.54pt]{figures/7.jpg}};
\draw  [fill={rgb, 255:red, 155; green, 155; blue, 155 }  ,fill opacity=0.55 ] (609.8,213) -- (646.1,223.89) -- (646.1,250.11) -- (609.8,261) -- cycle ;
\draw [color={rgb, 255:red, 53; green, 88; blue, 13 }  ,draw opacity=1 ][line width=1.5]    (573.87,236.27) -- (599.4,236.21) ;
\draw [shift={(603.4,236.2)}, rotate = 179.87] [fill={rgb, 255:red, 53; green, 88; blue, 13 }  ,fill opacity=1 ][line width=0.08]  [draw opacity=0] (11.61,-5.58) -- (0,0) -- (11.61,5.58) -- cycle    ;
\draw  [fill={rgb, 255:red, 155; green, 155; blue, 155 }  ,fill opacity=0.55 ] (351.4,213) -- (387.7,223.89) -- (387.7,250.11) -- (351.4,261) -- cycle ;
\draw [color={rgb, 255:red, 53; green, 88; blue, 13 }  ,draw opacity=1 ][line width=1.5]    (315.47,236.27) -- (341,236.21) ;
\draw [shift={(345,236.2)}, rotate = 179.87] [fill={rgb, 255:red, 53; green, 88; blue, 13 }  ,fill opacity=1 ][line width=0.08]  [draw opacity=0] (11.61,-5.58) -- (0,0) -- (11.61,5.58) -- cycle    ;
\draw  [fill={rgb, 255:red, 155; green, 155; blue, 155 }  ,fill opacity=0.55 ] (177.4,28.2) -- (213.7,39.09) -- (213.7,65.31) -- (177.4,76.2) -- cycle ;
\draw  [fill={rgb, 255:red, 155; green, 155; blue, 155 }  ,fill opacity=0.55 ] (227.8,27.4) -- (264.1,38.29) -- (264.1,64.51) -- (227.8,75.4) -- cycle ;
\draw  [fill={rgb, 255:red, 155; green, 155; blue, 155 }  ,fill opacity=0.55 ] (307.8,27.4) -- (344.1,38.29) -- (344.1,64.51) -- (307.8,75.4) -- cycle ;
\draw  [fill={rgb, 255:red, 155; green, 155; blue, 155 }  ,fill opacity=0.55 ] (438.2,27.8) -- (474.5,38.69) -- (474.5,64.91) -- (438.2,75.8) -- cycle ;
\draw [color={rgb, 255:red, 132; green, 28; blue, 153 }  ,draw opacity=1 ][line width=1.5]    (429,51.34) -- (349.8,50.2) ;
\draw [shift={(433,51.4)}, rotate = 180.83] [fill={rgb, 255:red, 132; green, 28; blue, 153 }  ,fill opacity=1 ][line width=0.08]  [draw opacity=0] (11.61,-5.58) -- (0,0) -- (11.61,5.58) -- cycle    ;

\draw (148.31,230.89) node [anchor=north west][inner sep=0.75pt]  [font=\small] [align=left] {\begin{minipage}[lt]{20.4pt}\setlength\topsep{0pt}
\begin{center}
DNN
\end{center}

\end{minipage}};
\draw (77,303) node [anchor=north west][inner sep=0.75pt]  [font=\small] [align=left] {Client 1};
\draw (281,302) node [anchor=north west][inner sep=0.75pt] [font=\small]  [align=left] {Client 2};
\draw (546,301) node [anchor=north west][inner sep=0.75pt] [font=\small]  [align=left] {Client K};
\draw (415,230) node [anchor=north west][inner sep=0.75pt] [font=\Huge]  [align=left] {\ldots};
\draw (115.2,41.8) node [anchor=north west][inner sep=0.75pt]   [font=\small] [align=left] {\begin{minipage}[lt]{38.44pt}\setlength\topsep{0pt}
\begin{center}
Central \\Server
\end{center}

\end{minipage}};
\draw (613.11,231.29) node [anchor=north west][inner sep=0.75pt]  [font=\small] [align=left] {\begin{minipage}[lt]{20.4pt}\setlength\topsep{0pt}
\begin{center}
DNN
\end{center}

\end{minipage}};
\draw (354.71,231.29) node [anchor=north west][inner sep=0.75pt]  [font=\small] [align=left] {\begin{minipage}[lt]{20.4pt}\setlength\topsep{0pt}
\begin{center}
DNN
\end{center}

\end{minipage}};
\draw (180.71,44.89) node [anchor=north west][inner sep=0.75pt]  [font=\small] [align=left] {\begin{minipage}[lt]{20.4pt}\setlength\topsep{0pt}
\begin{center}
DNN
\end{center}

\end{minipage}};
\draw (231.11,45.29) node [anchor=north west][inner sep=0.75pt]  [font=\small] [align=left] {\begin{minipage}[lt]{20.4pt}\setlength\topsep{0pt}
\begin{center}
DNN
\end{center}

\end{minipage}};
\draw (311.91,45.29) node [anchor=north west][inner sep=0.75pt]  [font=\small] [align=left] {\begin{minipage}[lt]{20.4pt}\setlength\topsep{0pt}
\begin{center}
DNN
\end{center}

\end{minipage}};
\draw (268.5,47.29) node [anchor=north west][inner sep=0.75pt] [font=\huge]  [align=left] {\ldots};
\draw (442.31,44.89) node [anchor=north west][inner sep=0.75pt]  [font=\small] [align=left] {\begin{minipage}[lt]{20.4pt}\setlength\topsep{0pt}
\begin{center}
DNN
\end{center}

\end{minipage}};
\draw (130,163.4) node [anchor=north west][inner sep=0.75pt]  [font=\normalsize]  {$w_{1}$};
\draw (264.6,163.4) node [anchor=north west][inner sep=0.75pt]  [font=\normalsize]  {$w_{2}$};
\draw (480.6,172) node [anchor=north west][inner sep=0.75pt] [font=\normalsize]   {$w_{K}$};
\draw (446.8,73.4) node [anchor=north west][inner sep=0.75pt]  [font=\normalsize]  {$w^{*}$};
\draw [color={rgb, 255:red, 53; green, 88; blue, 13 }] (110,213) node [anchor=north west][inner sep=0.75pt]   [align=left] {\begin{minipage}[lt]{8.67pt}\setlength\topsep{0pt}
\begin{center}
\numcircled{1}
\end{center}

\end{minipage}};
\draw [color={rgb, 255:red, 53; green, 88; blue, 13 }] (315.2,213) node [anchor=north west][inner sep=0.75pt]   [align=left] {\begin{minipage}[lt]{8.67pt}\setlength\topsep{0pt}
\begin{center}
\numcircled{1}
\end{center}

\end{minipage}};
\draw [color={rgb, 255:red, 53; green, 88; blue, 13 }] (573,213.4) node [anchor=north west][inner sep=0.75pt]   [align=left] {\begin{minipage}[lt]{8.67pt}\setlength\topsep{0pt}
\begin{center}
\numcircled{1}
\end{center}

\end{minipage}};
\draw [color={rgb, 255:red, 132; green, 28; blue, 153 }] (380.6,28) node [anchor=north west][inner sep=0.75pt]   [align=left] {\begin{minipage}[lt]{8.67pt}\setlength\topsep{0pt}
\begin{center}
\numcircled{3}
\end{center}

\end{minipage}};
\draw [color={rgb, 255:red, 134; green, 24; blue, 24 }] (148.6,122.2) node [anchor=north west][inner sep=0.75pt]   [align=left] {\begin{minipage}[lt]{8.67pt}\setlength\topsep{0pt}
\begin{center}
\numcircled{2}
\end{center}

\end{minipage}};
\draw [color={rgb, 255:red, 134; green, 24; blue, 24 }] (252.8,122.2) node [anchor=north west][inner sep=0.75pt]   [align=left] {\begin{minipage}[lt]{8.67pt}\setlength\topsep{0pt}
\begin{center}
\numcircled{2}
\end{center}

\end{minipage}};
\draw [color={rgb, 255:red, 134; green, 24; blue, 24 }] (405.6,121.8) node [anchor=north west][inner sep=0.75pt]   [align=left] {\begin{minipage}[lt]{8.67pt}\setlength\topsep{0pt}
\begin{center}
\numcircled{2}
\end{center}

\end{minipage}};
\draw [color={rgb, 255:red, 28; green, 74; blue, 128 }] (210.2,160.6) node [anchor=north west][inner sep=0.75pt]   [align=left] {\begin{minipage}[lt]{8.67pt}\setlength\topsep{0pt}
\begin{center}
\numcircled{4}
\end{center}

\end{minipage}};
\draw [color={rgb, 255:red, 28; green, 74; blue, 128 }] (340,143.6) node [anchor=north west][inner sep=0.75pt]   [align=left] {\begin{minipage}[lt]{8.67pt}\setlength\topsep{0pt}
\begin{center}
\numcircled{4}
\end{center}

\end{minipage}};
\draw [color={rgb, 255:red, 28; green, 74; blue, 128 }] (505.8,133.6) node [anchor=north west][inner sep=0.75pt]   [align=left] {\begin{minipage}[lt]{8.67pt}\setlength\topsep{0pt}
\begin{center}
\numcircled{4}
\end{center}

\end{minipage}};

\end{tikzpicture}

%% file: nb_clients_comp.tex
\begin{table*}[!htp]
\renewcommand{\arraystretch}{1.1}
\setlength\tabcolsep{12pt}
\caption{$F_1$ Scores (\%) Obtained by the Selected FL Algorithms Under\\Different Number $K$ of Clients and the Decentralization Scenarios}
\label{tab:DS_33}
\centering
\begin{tabular}{@{}llccccccccccc@{}}
\hline
\multirow{2}{*}{Algorithm} & & \multicolumn{3}{c}{$K$=7} & & \multicolumn{3}{c}{$K$=14} &                                      & \multicolumn{3}{c}{$K$=28}                           \\ \cline{3-5}\cline{7-9}\cline{11-13} 
             & & {DS1} & {DS2} & {DS3} & & {DS1} & {DS2} & {DS3} & & {DS1} & {DS2} & DS3 \\ \hline
FedAvg \cite{mcmahan2017communication} &         & {79.2}    & {52.2}    & {47.5}  &  & {79.6}    & {51.3}    & {48.7}  &  & {77.3}    & {48.5}    & 47.6    \\ \hline
\multirow{4}{*}{\thead[l]{Local\\Training\\Focused}} & FedProx \cite{li2020federated}         & {79.5}    & {56.3}    & {53.3}  &  & {79.7}    & {56.2}    & {55.6}  &  & {77.4}    & {57.8}    & 54.5    \\ 
& SCAFFOLD \cite{karimireddy2020scaffold}        & {78.6}    & {57.2}    & {54.4}  &  & {80.0}    & {56.9}    & {53.3}  &  & {78.7}      & {54.2}      &   51.3   \\ 
& MOON \cite{li2021model}          & \textbf{81.7}    & {59.5}    & {55.2}  &  & {81.2}    & {56.5}    & {54.7}  &  & {80.1}    & {55.3}    & \textbf{54.9}    \\ 
& FedDC \cite{gao2022feddc}          & {78.8}    & {58.4}    & \textbf{57.4}  &  & {81.3}    & {57.2}    & {53.9} &   & \textbf{80.2}      & {55.4}      &   54.1    \\ \hline
\multirow{3}{*}{\thead[l]{Model\\Aggregation\\Focused}} & FedNova \cite{wang2021novel}         & {80.6}    & {53.7}    & {48.6} &   & {80.4}    & {52.2}    & {47.7}  &  & {79.4}    & {50.4}    & 46.1    \\ 
& pFedLA \cite{ma2022layer}         & {81.3}    & {56.8}    & {53.6}  &  & \textbf{81.4}      & {55.9}      & {53.2}   &   & {79.5}      & {54.1}      &   50.9    \\ 
& FedBN \cite{li2021fedbn}          & {78.4}    & \textbf{62.3}    & {57.1}  &  & {78.3}      & \textbf{61.2}      & \textbf{57.2}   &   & {77.2}      & \textbf{60.4}      &   56.3    \\ \hline
\end{tabular}
\end{table*}

%% file: sensitivity_analysis.tex
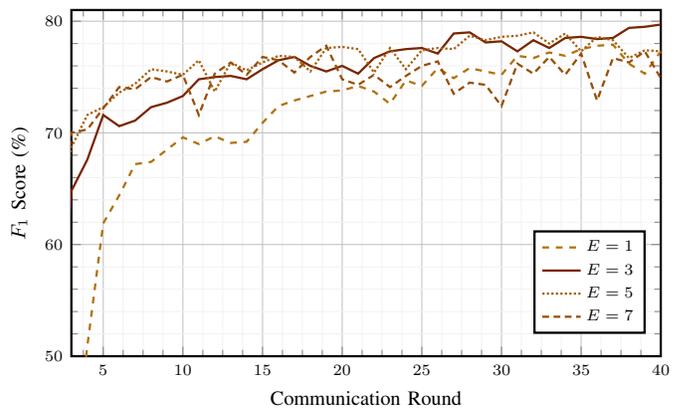
\begin{figure}[t]
    \centering
    \begin{tikzpicture}[scale = 0.85]%
	\begin{axis}[
    legend columns=1,
    height=7cm,
    width=10.8cm,
    legend style={font=\scriptsize, at={(axis cs:32,52)},anchor=south west},
    grid=both,
    grid style={line width=.1pt, draw=gray!10},
    major grid style={line width=.2pt,draw=gray!50},
    minor x tick num=3,
    minor y tick num=4,
    xlabel= {\small Communication Round},
    ylabel= {\small $F_1$ Score (\%)},
    xmin=3,xmax=40,
    ymin=50, ymax=81]
    \addplot+[name path=capacity,dashed,color=FedProx_c!10!yellow!50!Brown,mark options={fill=green!70!white},line width=1pt] table [x=comm_rounds, y=local-epoch-1, col sep=comma] {sensitivity_analysis.csv};\addlegendentry{$E=1$};
    \addplot+[name path=capacity,solid,color=FedProx_c,mark options={fill=white},line width=1pt] table [x=comm_rounds, y=local-epoch-3, col sep=comma] {sensitivity_analysis.csv};\addlegendentry{$E=3$};
    \addplot+[name path=capacity,densely dotted,color=FedProx_c!30!yellow!50!Brown,mark options={fill=white},line width=1pt] table [x=comm_rounds, y=local-epoch-5, col sep=comma] {sensitivity_analysis.csv};\addlegendentry{$E=5$}
    \addplot+[name path=capacity,densely dashed,color=FedProx_c!50!yellow!50!Brown,mark options={fill=white},line width=1pt] table [x=comm_rounds, y=local-epoch-7, col sep=comma] {sensitivity_analysis.csv};\addlegendentry{$E=7$};
	\end{axis}
    \end{tikzpicture}
\caption{$F_1$ score versus communication round obtained by the FedProx algorithm when the different numbers $E$ of local epochs are considered under the DS1.}
\label{fig:sens}
\end{figure}

%% file: learning_efficiency.tex
\begin{figure}[t]
    \centering
    \begin{tikzpicture}[scale = 0.85]%
	\begin{axis}[
    legend columns=1,
    height=7cm,
    legend cell align={left},
    width=10.8cm,
    legend style={font=\scriptsize, at={(axis cs:7.3,42)},anchor=south west},
    grid=both,
    grid style={line width=.1pt, draw=gray!10},
    major grid style={line width=.2pt,draw=gray!50},
    minor x tick num=3,
    minor y tick num=4,
    xlabel= {\small Communication Round},
    ylabel= {\small $F_1$ Score (\%)},
    xmin=1.5,xmax=10,
    ymin=40, ymax=78]
    \addplot+[name path=capacity,solid,color=SCAFFOLD_c,mark=*,mark options={fill=white},line width=1pt] table [x=comm_rounds, y=scaffold, col sep=comma] {learning_efficiency.csv};\addlegendentry{SCAFFOLD \cite{karimireddy2020scaffold}};
    \addplot+[name path=capacity,solid,color=FedAvg_c,mark=+,mark options={fill=white},line width=1pt] table [x=comm_rounds, y=fedavg, col sep=comma] {learning_efficiency.csv};\addlegendentry{FedAvg \cite{mcmahan2017communication}};
    \addplot+[name path=capacity,solid,color=MOON_c,mark=asterisk,mark options={fill=white},line width=1pt] table [x=comm_rounds, y=moon, col sep=comma] {learning_efficiency.csv};\addlegendentry{MOON \cite{li2021model}}
    \addplot+[name path=capacity,solid,color=FedDC_c,mark=oplus,mark options={fill=white},line width=1pt] table [x=comm_rounds, y=feddc, col sep=comma] {learning_efficiency.csv};\addlegendentry{FedDC \cite{gao2022feddc}};
    \addplot+[name path=capacity,solid,color=FedNova_c,mark=square,mark options={fill=white},line width=1pt] table [x=comm_rounds, y=fednova, col sep=comma] {learning_efficiency.csv};\addlegendentry{FedNova \cite{wang2021novel}};
    \addplot+[name path=capacity,solid,color=FedProx_c,mark=triangle,mark options={fill=white},line width=1pt] table [x=comm_rounds, y=fedprox, col sep=comma] {learning_efficiency.csv};\addlegendentry{FedProx \cite{li2020federated}};
    \addplot+[name path=capacity,solid,color=pFedLA_c,mark=diamond,mark options={fill=white},line width=1pt] table [x=comm_rounds, y=pfedla, col sep=comma] {learning_efficiency.csv};\addlegendentry{pFedLA \cite{ma2022layer}};
    \addplot+[name path=capacity,solid,color=FedBN_c,mark=pentagon,mark options={fill=white},line width=1pt] table [x=comm_rounds, y=fedbn, col sep=comma] {learning_efficiency.csv};\addlegendentry{FedBN \cite{li2021fedbn}};
	\end{axis}
    \end{tikzpicture}
\caption{$F_1$ score versus communication round obtained by the investigated FL algorithms under the DS1.} 
\label{fig:le}
\end{figure}
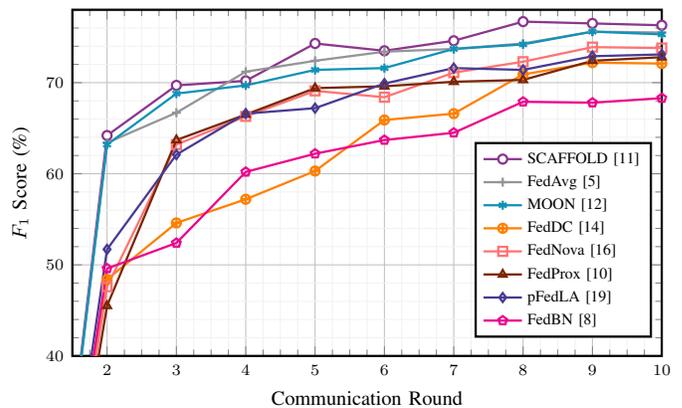